\documentclass[sn-mathphys-num,iicol]{sn-jnl}

\usepackage{graphicx}%
\usepackage{multirow}%
\usepackage{amsmath,amssymb,amsfonts}%
\usepackage{amsthm}%
\usepackage{mathrsfs}%
\usepackage[title]{appendix}%
\usepackage[table,xcdraw,dvipsnames]{xcolor}
\usepackage{textcomp}%
\usepackage{manyfoot}%
\usepackage{booktabs}%
\usepackage{algorithm}%
\usepackage{algorithmicx}%
\usepackage{algpseudocode}%
\usepackage{listings}%
\usepackage{bm}
\usepackage{bbm}
\usepackage{threeparttable}
\usepackage{wrapfig}
\usepackage{caption}

\DeclareMathOperator*{\argmin}{arg\,min}

\theoremstyle{thmstyleone}%
\theoremstyle{thmstyletwo}%

\theoremstyle{thmstylethree}%

\newcommand{\oldnew}[2]{#2}
\newcommand{\boldparagraph}[1]{\vspace{0.2cm}\noindent{\bf #1}}

\newcommand{\etc}{\textit{etc}.}
\newcommand{\ie}{\textit{i}.\textit{e}.}
\newcommand{\eg}{\textit{e}.\textit{g}.}
\begin{document}

\title[Article Title]{MapTRv2: An End-to-End Framework for Online Vectorized HD Map Construction}


\author[1,2]{\fnm{Bencheng} \sur{Liao}}\email{bcliao@hust.edu.cn}
\author[2]{\fnm{Shaoyu} \sur{Chen}}\email{shaoyuchen@hust.edu.cn}
\author[2]{\fnm{Yunchi} \sur{Zhang}}\email{zyc10ud@hust.edu.cn}
\author[2]{\fnm{Bo} \sur{Jiang}}\email{bjiang@hust.edu.cn}
\author[3]{\fnm{Qian} \sur{Zhang}}\email{qian01.zhang@horizon.ai}
\author[2]{\fnm{Wenyu} \sur{Liu}}\email{liuwy@hust.edu.cn}
\author[3]{\fnm{Chang} \sur{Huang}}\email{chang.huang@horizon.ai}
\author*[2]{\fnm{Xinggang} \sur{Wang}}\email{xgwang@hust.edu.cn}
\affil[1]{\orgdiv{Institute of Artificial Intelligence}, \orgname{Huazhong University of Science and Technology},  \orgaddress{\city{Wuhan}, \country{China}}}

\affil[2]{\orgdiv{School of Electronic Information and Communications}, \orgname{Huazhong University of Science and Technology}, \orgaddress{\city{Wuhan}, \country{China}}}

\affil[3]{\orgname{Horizon Robotics}, \orgaddress{\city{Beijing}, \country{China}}}


\abstract{High-definition (HD) map provides abundant and precise static environmental information of the driving scene, serving as a fundamental and indispensable component for planning in autonomous driving system.
In this paper, we present \textbf{Map} \textbf{TR}ansformer, an end-to-end framework for online vectorized HD map construction.
We propose a unified permutation-equivalent modeling approach,
\ie, modeling map element as a point set with a group of equivalent permutations, which accurately describes the shape of map element and stabilizes the learning process. We design a hierarchical query embedding scheme to flexibly encode structured map information and perform hierarchical bipartite matching for map element learning. 
To speed up convergence, we further introduce auxiliary one-to-many matching and dense supervision.
The proposed method well copes with various map elements with arbitrary shapes.
It runs at real-time inference speed and achieves state-of-the-art performance on both nuScenes and Argoverse2 datasets.
Abundant qualitative results show  stable and robust map construction quality in complex and various driving scenes. 
Code and more demos are available at \url{https://github.com/hustvl/MapTR} for facilitating further studies and applications.}




\maketitle
\section{Introduction}\label{sec1}
High-definition   (HD) map is the high-precision  map specifically designed for autonomous driving, composed of instance-level vectorized representation of  map elements (pedestrian crossing, lane divider, road boundaries, centerline, \etc). HD map contains rich semantic information about road topology and traffic rules, which is essential for the navigation of self-driving vehicles.

Conventionally HD map is constructed offline using SLAM-based methods~\cite{loam,legoloam,liosam}, which brings many issues:
1) complicated pipeline and high cost;
2) difficult to keep maps up-to-date;
3) misalignment with ego vehicle and high localization error (empirically, 0.4m longitudinally and 0.2m laterally).

Given these limitations, recently, online HD map construction has attracted ever-increasing interest, which constructs map around ego-vehicle at runtime with vehicle-mounted sensors and well solves the problems above.

\begin{figure}[t!]
    \centering
    \includegraphics[width=0.98\linewidth]{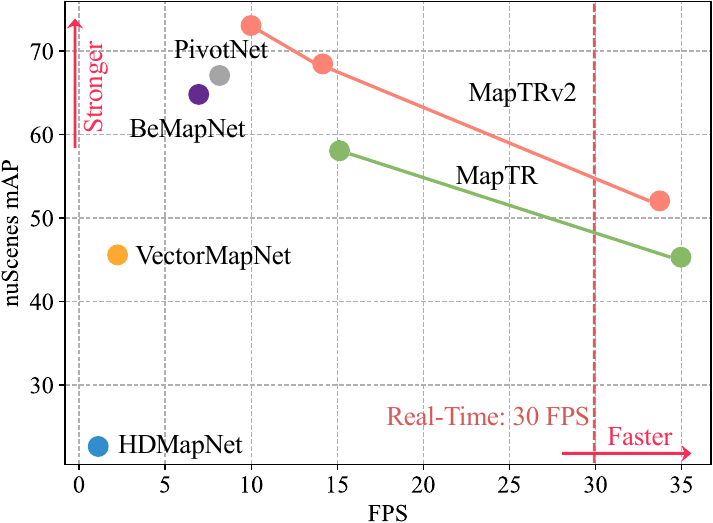}
    \vspace*{-0.1cm}
    \caption{Speed-accuracy trade-off comparisons.
    The proposed MapTRv2 outperforms previous state-of-the-art methods in terms of both speed (FPS) and accuracy (mAP). Compared with MapTR, MapTRv2 further improves performance by a large margin. The FPSs are measured on one NVIDIA RTX 3090. }
    \label{fig:tradeoff}
    \vspace*{-0.5cm}
\end{figure}

Early works~\cite{chen2022persformer,lstr,can2021structured} leverage line-shape priors to perceive open-shape lanes only on the front-view image. They are restricted to single-view perception and can not cope with other map elements with arbitrary shapes.
With the development of bird's eye view (BEV) representation learning,
recent works~\cite{gkt,cvt,fiery,bevformer} predict rasterized map by performing BEV semantic segmentation. 

However, the rasterized map lacks vectorized instance-level information, such as the lane structure,  which is important for the downstream tasks (\eg, motion prediction and planning). 
To construct the vectorized HD map, HDMapNet~\cite{hdmapnet} groups pixel-wise segmentation results into vectorized instances, which requires complicated and time-consuming post-processing.   VectorMapNet~\cite{vectormapnet} represents each map element as a point sequence. It adopts a cascaded coarse-to-fine framework  and utilizes an auto-regressive decoder to predict points sequentially, leading to long inference time and error accumulation.

Current online vectorized HD map construction methods are restricted by the efficiency and precision, especially in real-time scenarios.
Recently, DETR~\cite{detr} employs a simple and efficient encoder-decoder Transformer architecture and achieves end-to-end object detection.  
It is natural to ask a question: \textit{Can we design a DETR-like paradigm for efficient end-to-end vectorized HD map construction?} We show that the answer is affirmative with our proposed  \textbf{Map} \textbf{TR}ansformer.

In our previous conference version~\cite{maptr} MapTR, we propose a novel permutation-equivalent modeling for map elements that do not have fixed orderings, such as lane dividers, road boundaries, and pedestrian crossings.
This approach models each map element as a set of points, accompanied by a group of equivalent permutations. The set of points defines the position of the map element, while the group of permutations includes all possible organizational sequences of the points that correspond to the same geometric shape, thereby eliminating ambiguity in shape representation. Building on this permutation-equivalent modeling, we developed a Transformer-based encoder-decoder framework named MapTR. This framework processes data from vehicle-mounted sensors to output vectorized HD maps. We streamline the process of constructing online vectorized HD maps into a parallel regression problem and introduced hierarchical query embeddings to efficiently encode information at both the instance and point levels.

Though MapTR has become a popular state-of-the-art (SOTA) method in the field of online HD map construction and has attracted widespread attention. There still lie issues in MapTR. Initially, MapTR is designed to model map elements lacking physical directions. 
Yet, the lane centerline, a crucial element for downstream motion prediction and planning, inherently possesses practical direction and connection, which MapTR does not account for.
Furthermore, MapTR flattens the hierarchical query embeddings to formulate very long sequence and directly applies vanilla self-attention on the sequence. Such implementation results in huge memory and computation cost,  prohibiting scaling MapTR with more instances and more points.

To address the aforementioned issues, we introduce MapTRv2, an advancement over the conference version.
MapTRv2 treats the lane centerline graph as paths~\cite{lanegap} through a graph traversing algorithm and proposes semantic-aware shape modeling and matching. 
Specifically, for map elements of centerline semantic, MapTRv2 does not permute them and directly chooses the given order as targeted permutation.
To reduce the cost of computation and memory, decoupled self-attention mechanism is proposed for interaction among queries, \ie, respectively performing attention along the inter-instance dimension and intra-instance dimension. Compared to the vanilla self-attention used in MapTR, MapTRv2 decreases the memory consumption and computation complexity from $O((N\times N_v)^2)$ to $O(N^2+N_{v}^2)$, where $N$ and $N_v$ are respectively the number of instance queries and point queries.

With above improvements, we can use the same framework to simultaneously predict both undirected map elements and directed map elements with lower computation complexity and memory consumption. To further speed up the convergence, we also advance the training recipe of MapTR. we add an auxiliary one-to-many matching branch during training, which increases the ratio of positive samples.
To  further leverage semantic and geometric information, we introduce auxiliary foreground segmentation on both perspective view (PV) and bird's eye view (BEV), and leverage depth supervision to guide the backbone to learn 3D geometrical information.

We conduct extensive experiments on both nuScenes~\cite{nuscenes} dataset and Argoverse2~\cite{av2} dataset. The results show that MapTRv2 significantly ourperforms MapTR and previous SOTA methods in terms of both accuracy and efficiency. In particular, on nuScenes, MapTRv2 based on ResNet18 backbone runs at real-time inference speed (33.7 FPS) on RTX 3090, $15\times$ faster than the previous state-of-the-art (SOTA) camera-based method VectorMapNet while achieving 6.3 higher mAP.
With ResNet50 as the backbone,  MapTRv2 achieves 68.7 mAP at 14.1 FPS, which is 22.7 mAP higher and $6\times$ faster than VectorMapNet-ResNet50, and even outperforms multi-modality VectorMapNet. MapTRv2 also demonstrates faster convergence than MapTR-ResNet50 (4$\times$ shorter training schedule and 2.8 higher mAP, see Fig.~\ref{fig:convergence}). 
When scaling up to a larger backbone VoVNetV2-99, MapTRv2 sets a new record (73.4 mAP and 9.9 FPS) with only-camera input.
On Argoverse2, MapTRv2 outperforms VectorMapNet by 28.9 mAP in terms of 3D map construction with the same backbone ResNet50.
As the visualization shows (Fig.~\ref{fig:av2} and Fig.~\ref{fig:nusc}), MapTRv2 maintains stable and robust map construction quality in complex and various driving scenes.

Our contributions can be summarized as follows:
\begin{itemize}
    \item We propose a unified permutation-based modeling approach for all kinds of map elements (directed and undirected), \ie, modeling map element as a point set with a group of permutations, which accurately describes the shape of map element and stabilizes the learning process.
    \item  Based on the novel modeling, we present a structured end-to-end framework for efficient online vectorized HD map construction.
    We design a hierarchical query embedding scheme to flexibly encode instance-level and point-level information and propose decoupled self-attention to reduce the memory consumption and computation complexity.

    \item  We introduce advanced training recipe targeted for the simple encoder-decoder framework, largely boosting the performance and improving the convergence.
    \item The proposed MapTRv2 achieves best accuracy and speed tradeoff at different model sizes in a fully end-to-end manner.
\end{itemize}

This journal paper (MapTRv2) is an extension of a conference paper (MapTR) published at ICLR 2023~\cite{maptr}.
Improvements compared to the former version are as follows. First, 
MapTRv2 extends MapTR to modeling and learning directed centerline, which is important for downstream motion planning~\cite{nuplan_garage,gu2024producing}.
Second, we introduce decoupled self-attention tailored for the hierarchical query mechanism, which greatly reduces the memory consumption and brings gain.
Third, we adopt auxiliary dense supervision on both perspective view and bird's eyes view, which significantly boosts the performance.
Fourth, we introduce  auxiliary one-to-many set prediction branch to speed up the convergence.
Fifth, we provide more theoretical analysis and discussions about the proposed modules, which reveal more about the working mechanism of our framework. 
Finally, we extend the framework to 3D map construction (the conference version learns 2D map),
and provide additional experiments on Argoverse2 dataset~\cite{av2}.

\section{Related Work}\label{sec2}

\begin{figure*}[htbp!]
    \centering
    \includegraphics[width=0.98\linewidth]{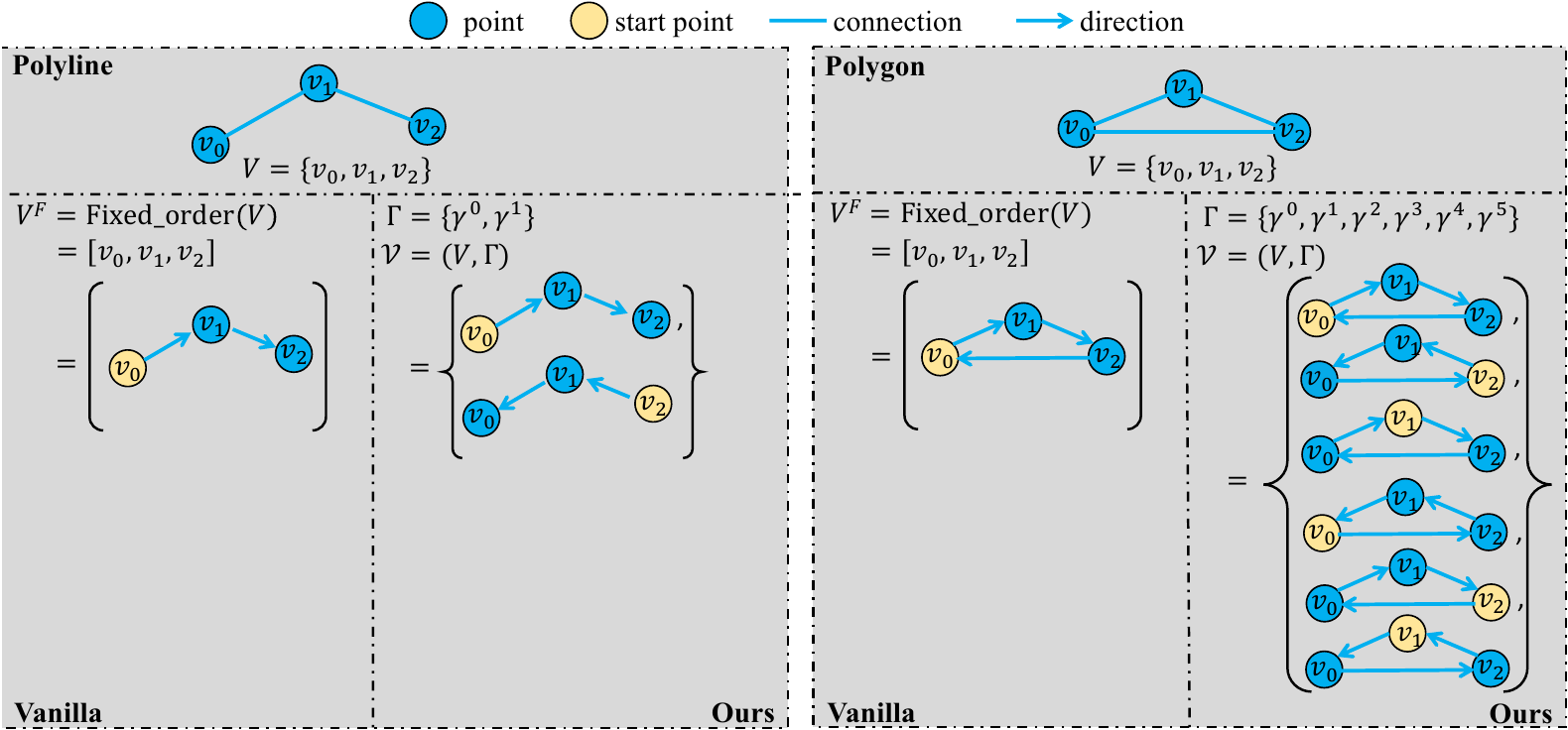}
    \vspace*{-0.1cm}
    \caption{Illustration of permutation-equivalent shape modeling. Map elements are geometrically abstracted and discretized into polylines and polygons.
    MapTRv2 models each map element  with $(V, \Gamma)$ (a point set $V$ and a group of equivalent permutations $\Gamma$), avoiding ambiguity and stabilizing the learning process. A special case is, if the polyline element has a specific direction (\eg, centerline), $\Gamma$ includes only one permutation.}
    \label{fig:modeling}
    \vspace*{-0.5cm}
\end{figure*}

\boldparagraph{HD Map Construction.}
Recently, with the development of PV-to-BEV methods~\cite{Ma2022VisionCentricBP, li2022delving},
HD map construction is formulated as a segmentation problem based on  surround-view image data captured by vehicle-mounted cameras. \cite{gkt,cvt,fiery,bevformer,lss,polarbev,liu2022bevfusion,baeformer,uniformer,beverse} generate rasterized map by performing BEV semantic segmentation.
To build vectorized HD map, HDMapNet~\cite{hdmapnet}  groups pixel-wise semantic segmentation results with heuristic and time-consuming post-processing to generate vectorized instances. 
VectorMapNet~\cite{vectormapnet} serves as the first end-to-end framework, which adopts a two-stage coarse-to-fine framework and utilizes auto-regressive decoder to predict points sequentially, leading to long inference time and the ambiguity about permutation.
Different from VectorMapNet, our method introduces novel and unified shape modeling for map elements, solving the ambiguity and stabilizing the learning process. And we further build a structured and parallel one-stage framework with much higher efficiency. Concurrent and follow-up works of our conference version~\cite{maptr} focus on other promising designs of end-to-end HD map construction~\cite{shin2023instagram, qiao2023end, zhang2023online, chen2023polydiffuse, qiao2023machmap,li2023topology,yuan2024streammapnet,xu2023insightmapper,li2024lanesegnet,yuan2024presight, chen2024maptracker,jiang2024p,wang2024stream,xiong2024ean,kalfaoglu2023topomask,liu2024leveraging} and extend to other relative tasks~\cite{nmp,toponet,lanegap,chen2023vma,pip,vad}. 
BeMapNet~\cite{qiao2023end} and PivotNet~\cite{pivotnet} propose  Bezier-based representation and pivot-based representation for modeling map geometry. 
NMP~\cite{nmp} introduces a neural representation of global maps that enables automatic global map updates and enhances local map inference performance. 
TopoNet~\cite{toponet} proposes a comprehensive framework to learn the connection relationship of lanes and the assignment relationship between lanes and traffic elements.
PolyDiffuse~\cite{chen2023polydiffuse} introduces diffusion mechanism into MapTR to further refine the results via a conditional generation procedure. MapVR~\cite{zhang2023online} applies differentiable rasterization to the vectorized results produced by MapTR to incorporate precise and geometry-aware supervision.
MapTRv2 extends MapTR to a more general framework. It supports centerline  learning and 3D map construction. It also achieves higher accuracy and faster convergence speed.

\boldparagraph{Lane Detection.}
Lane detection can be viewed as a sub task of HD map construction, which focuses on detecting lane elements in the road scenes. Since most datasets of lane detection only provide single view annotations and focus on open-shape elements, related methods~\cite{BEVLaneDet,chen2022persformer,tabelini2021keep,lstr,wang2022keypoint,feng2022rethinking,can2021structured,Guo2020GenLaneNetAG,efrat20203d,liu2022learning,liu2021condlanenet} are restricted to single view.
LaneATT~\cite{tabelini2021keep} utilizes an anchor-based deep lane detection model to achieve good trade-off between accuracy and efficiency. LSTR~\cite{lstr} adopts the Transformer architecture to directly output parameters of a lane shape model. GANet~\cite{wang2022keypoint} formulates lane detection as a keypoint estimation and association problem and takes a bottom-up design.  \cite{feng2022rethinking} proposes parametric Bezier curve-based method for lane detection. Instead of detecting lanes in the 2D image coordinate, \cite{garnett20193d} proposes 3D-LaneNet which performs 3D lane detection in BEV. STSU~\cite{can2021structured} represents lanes as a directed graph in BEV coordinates and adopts the curve-based Bezier method to predict lanes from the monocular camera image. Persformer~\cite{chen2022persformer} provides better BEV feature representation and optimizes anchor design to unify 2D and 3D lane detection simultaneously.
BEV-LaneDet~\cite{BEVLaneDet} presents a virtual camera to guarantee consistency and proposes an efficient spatial transformation pyramid module.

\boldparagraph{Contour-based 2D Instance Segmentation.}
Another line of related work is contour-based 2D instance segmentation~\cite{Zhu_2022_CVPR, polarmask, xu2019explicit,liu2021dance, e2ec,feng2023recurrent,Xie2021PolarMaskEP,Acuna2018EfficientIA,wei2020point}. These methods  reformulate 2D instance segmentation as object contour prediction task, where the object contour is a closed-shape polygon, and estimate the image coordinates of the contour vertices. CurveGCN~\cite{ling2019fast} utilizes Graph Convolution Networks to predict polygonal boundaries. 
PolarMask~\cite{polarmask} produces the instance contour through instance center classification and dense distance regression in a polar coordinate system. Deepsnake~\cite{peng2020deep} proposes a two-stage contour evolution process and designs circular convolution to exploit the features of the contour. BoundaryFormer~\cite{Lazarow_2022_CVPR} adopts advanced Transformer architecture and follows the two-stage paradigm to generate the polygon vertices based on the intermediate box results predicted by the first stage. SharpContour~\cite{Zhu_2022_CVPR} proposes an efficient and generic contour-based boundary refinement approach to iteratively deform the contour by updating offsets in a discrete manner.

\boldparagraph{Detection Transformers.}
As a pioneer work, DETR~\cite{detr} eliminates the need for hand-crafted components (\eg, anchor generation, ruled-based label assignment, non-maximum suppression post-processing.), and  built the first fully end-to-end object detector. It represents object boxes as a set of queries and directly adopts a unified Transformer encoder-decoder architecture to perform object box detection. Due to the effectiveness and simplicity of DETR and its varieties~\cite{deformdetr,yolos,zhang2023dino,Li2022DNDETRAD,liu2022dabdetr,wang2022anchor,Meng2021ConditionalDF,jia2023detrs}, their paradigms have been widely transferred in many complex tasks, such as 2D instance segmentation~\cite{queryinst,mask2former,li2023mask}, 2D pose estimation~\cite{li2021pose,Shi2022EndtoEndMP,Li2021TokenPoseLK}, 3D object detection~\cite{wang2022detr3d,bevformer,lin2022sparse4d,liu2022petr,mao20233d,wang2023multi,shi2023pv}. Directly inheriting the simplicity and end-to-end manner from DETR, we propose MapTRv2 to efficiently produce high-quality vectorized HD map online.

\begin{figure*}[t!]
    \begin{center}
    \includegraphics[width=0.95\linewidth]{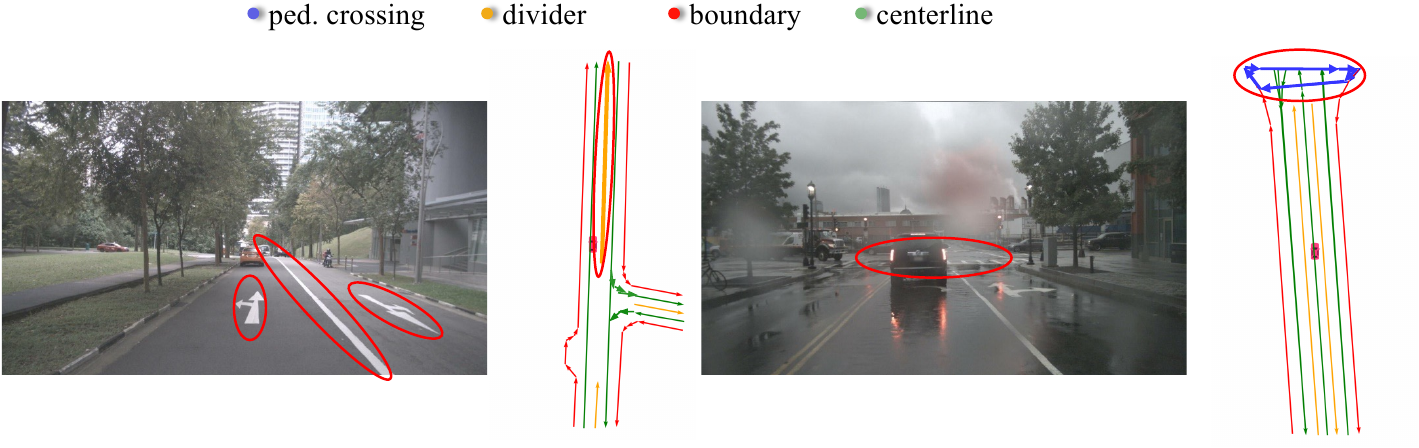}
    \end{center}
    \vspace*{-0.2cm}
    \caption{Typical cases for illustrating the ambiguity of map element in terms of start point and direction.
    Left: for polyline with unspecific direction (\eg, the lane divider between two opposite lanes), defining its direction is difficult. Both endpoints of the lane divider can be regarded as the start point and the point set can be organized in two directions.
    Right: for polygon (\eg, pedestrian crossing),
    each point of the polygon can be regarded as the start point, and the polygon can be connected in two opposite directions (counter-clockwise and clockwise). Note that some kinds of map elements (like centerline) have specific direction and have no ambiguity issue.}
    \label{fig:ambiguity-case}
    \vspace*{-0.4cm}
\end{figure*}
\section{Shape Modeling \label{sec:modeling}}
MapTRv2 aims at modeling and learning the HD map in a unified manner.
HD map is a collection of vectorized static map elements, including pedestrian crossing, lane divider, road boundary, centerline, \etc{}
For structured modeling,  MapTRv2 geometrically abstracts map elements as closed shape (like pedestrian crossing) and open shape (like lane divider). 
Through sampling points sequentially along the shape boundary, closed-shape element is discretized into polygon while open-shape element is discretized into polyline. 

Preliminarily,  both polygon and polyline can be represented as an ordered point set ${V}^{F}= [v_0, v_1, \dots, v_{N_v-1}]$ (see Fig.~\ref{fig:modeling} (Vanilla)). $N_v$ denotes the number of points. However, the permutation of the point set is not explicitly defined and not unique. There exist many equivalent permutations for polygon and polyline.
For example, as illustrated in Fig.~\ref{fig:ambiguity-case} (a), for the lane divider (polyline) between two opposite lanes, defining its direction is difficult. Both endpoints of the lane divider can be regarded as the start point and the point set can be organized in two directions.
In Fig.~\ref{fig:ambiguity-case} (b), for the pedestrian crossing (polygon), the point set can be organized in two opposite directions (counter-clockwise and clockwise). And circularly changing the permutation of point set has no influence on the geometrical shape of the polygon.
Imposing a fixed permutation to the point set as supervision is not rational. The imposed fixed permutation contradicts with other equivalent permutations, hampering the learning process.

To bridge this gap, MapTRv2 models each map element with $\mathcal{V}=(V,\Gamma)$. 
$V = \{v_{j}\}_{j=0}^{N_v-1}$ denotes the point set  of the map element ($N_v$ is the number of points). 
$\Gamma = \{\gamma^{k}\}$  denotes a group of equivalent permutations of the point set $V$, covering all the possible organization sequences.

Specifically, for polyline element with unspecific direction (see Fig.~\ref{fig:modeling} (left)),  $\Gamma$ includes $2$ kinds of equivalent permutations:
\begin{equation}
\begin{aligned}
\Gamma_{\rm{polyline}} = &\{\gamma^{0}, \gamma^{1} \} \\
 = &\begin{cases}
\gamma^{0}(j) = j \% N_v,\\
\gamma^{1}(j) = (N_v - 1) - j \% N_v.\\
\end{cases}
\end{aligned}
\end{equation}

For polyline element  with specific direction (\eg, centerline),  $\Gamma$ includes only one permutation:
$\{\gamma^0\}$.

For polygon element (see Fig.~\ref{fig:modeling} (right)),  $\Gamma$ includes $2\times N_v$ kinds of equivalent permutations:
\begin{equation}
\begin{aligned}
\Gamma_{\rm{polygon}} = &\{\gamma^{0}, \dots, \gamma^{2 \times N_v-1}\}\\
 = &\begin{cases}
\gamma^{0}(j)=j \% N_v,\\
\gamma^{1}(j) = (N_v - 1) - j \% N_v,\\
...\\
\gamma^{2 \times N_v - 2}(j) = (j + N_v - 1) \% N_v,\\
\gamma^{2 \times N_v - 1}(j) = (N_v - 1) - \\
\hspace{6.3em}(j + N_v - 1) \% N_v.
\end{cases}
\end{aligned}
\end{equation}

By introducing the conception of equivalent permutations, MapTRv2 models map elements in a unified manner and addresses the ambiguity issue. 
\begin{figure*}[t!]
    \centering
    \includegraphics[width=0.9\linewidth]{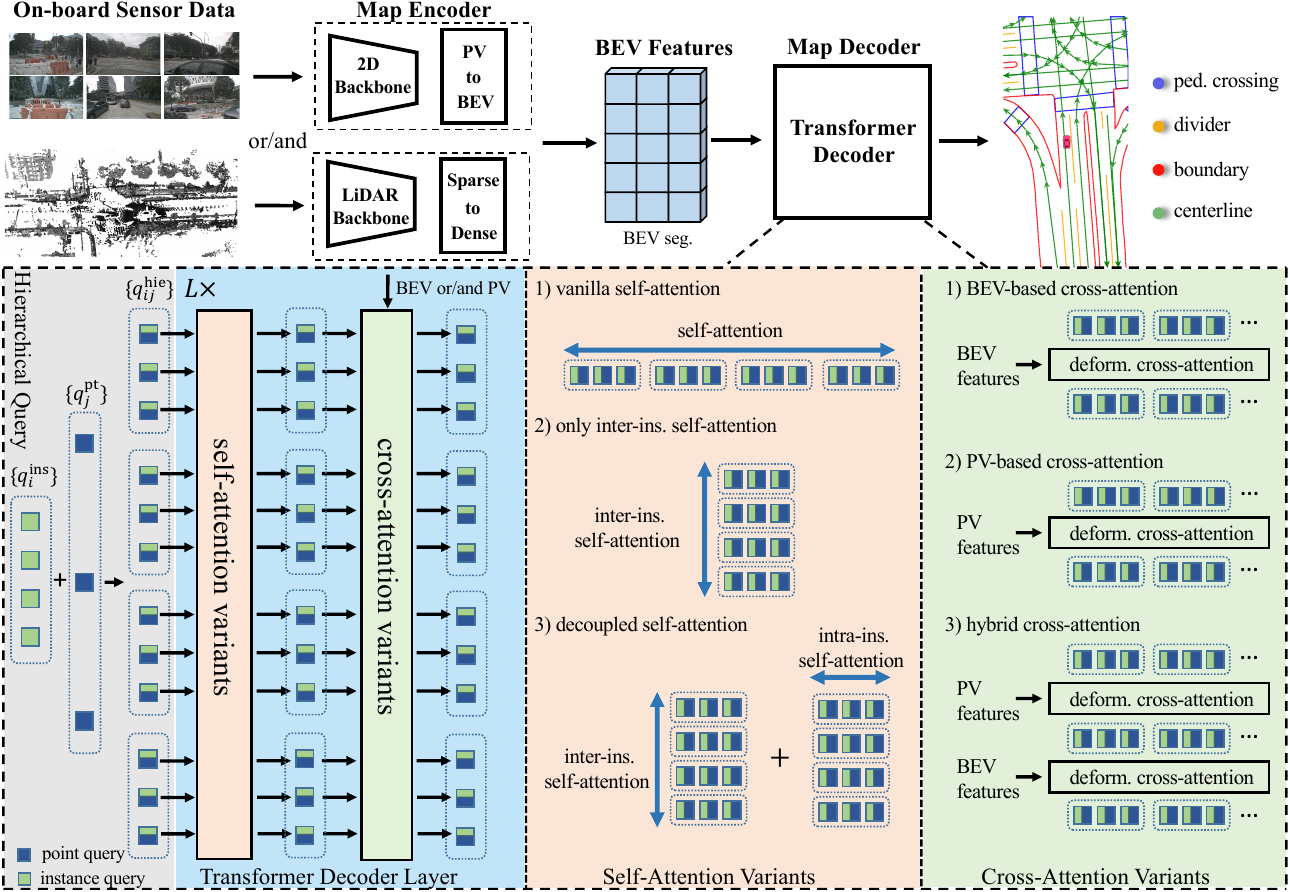}
    \caption{The overall architecture of MapTRv2. MapTRv2 adopts an encoder-decoder paradigm. The map encoder transforms sensor input to a unified BEV representation. The map decoder adopts a hierarchical query embedding scheme to explicitly encode  map elements. The $L$ stacked Transformer decoder layers iteratively refine the predicted map elements. We propose several self-attention variants and cross-attention variants to efficiently update the query features.  MapTRv2 is fully end-to-end. The pipeline is highly structured, compact and efficient.}
    \label{fig:framework}
\end{figure*}

\section{Architecture \label{sec:architecture}}
MapTRv2 adopts an encoder-decoder paradigm. The overall architecture is depicted in Fig.~\ref{fig:framework}.

\subsection{Map Encoder}
The map encoder  extracts features from sensor data and transforms the features into a unified feature representation, \ie, BEV representation.
MapTRv2 is compatible with various vehicle-mounted sensors.
Taking  multi-view images $\mathcal{I}=\{I_1, \ldots, I_M\} $ for example,
we leverage a conventional backbone to generate multi-view feature maps $\mathcal{F}=\{F_1, \ldots, F_M\} $.
Then PV image features $\mathcal{F}$ are transformed to BEV features  $\mathcal{B} \in \mathbb{R}^{H\times W \times C}$.
We support various PV2BEV transformation methods, \eg, CVT~\cite{cvt}, LSS~\cite{lss,liu2022bevfusion,bevdepth,bevdet}, Deformable Attention~\cite{bevformer,deformdetr}, GKT~\cite{gkt} and IPM~\cite{ipm}.
In MapTRv2, to explicitly exploit the depth information~\cite{bevdepth}, we choose LSS-based BEVPoolv2~\cite{bevpoolv2} as the default transformation method.  Extending MapTRv2 to multi-modality sensor data is straightforward and trivial.

\begin{figure*}[t!]
    \centering
    \includegraphics[width=0.9\linewidth]{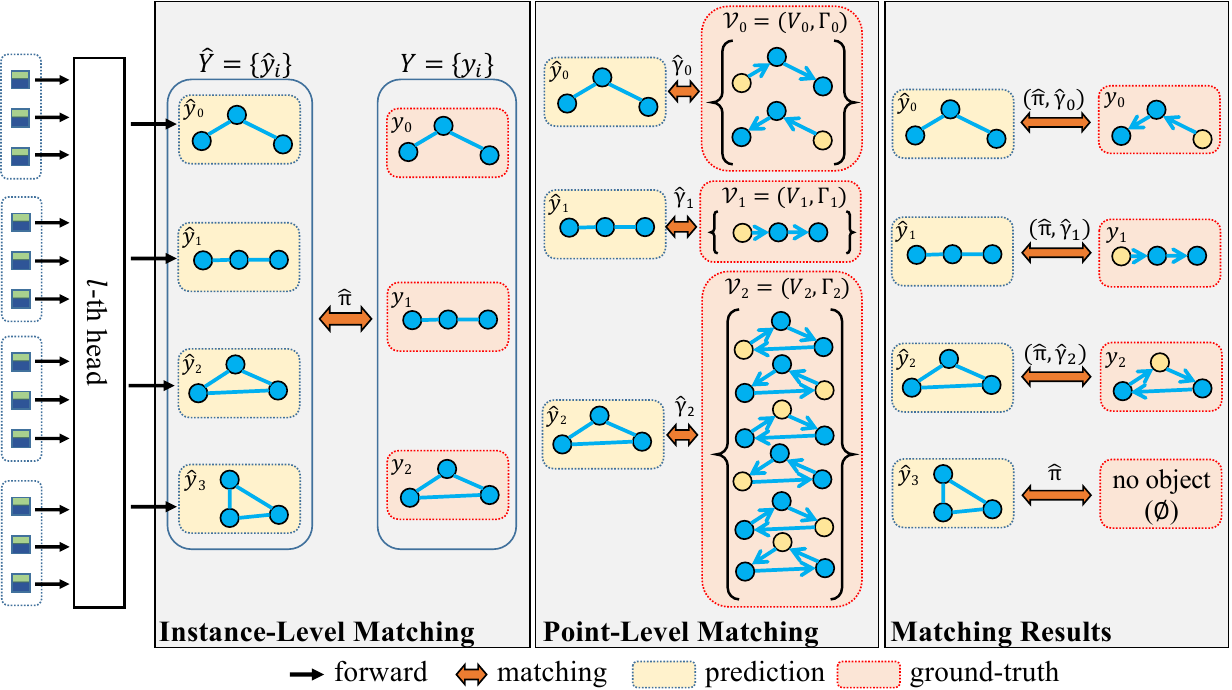}
    \caption{Hierarchical bipartite matching. MapTRv2 performs instance-level matching to find optimal instance-level assignment $\hat{\pi}$, and  performs point-level matching to find optimal point-to-point assignment $\hat{\gamma}$ (Sec.~\ref{sec:one2one_matching}). Based on the optimal instance-level and point-level assignments ($\hat{\pi}$ and $\{\hat{\gamma}_i\}$), one-to-one set prediction loss (Sec.~\ref{sec:one2one_loss}) is defined for end-to-end learning.}
    \label{fig:one2one}
    \vspace*{-0.3cm}
\end{figure*}
\subsection{Map Decoder}

Map decoder consists of map queries and several decoder layers. Each decoder layer utilizes self-attention and cross-attention to update the map queries. We detail the specific design as following:

\boldparagraph{Hierarchical Query.}
We propose a hierarchical query embedding scheme to explicitly encode each map element. Specifically, we define a set of instance-level queries $\{ q^{\rm{ins}}_i\}_{i=0}^{N-1}$ and a set of point-level queries $\{q^{\rm{pt}}_j\}_{j=0}^{N_v-1}$ shared by all instances. 
Each map element (with index $i$) corresponds to a set of hierarchical queries $\{q^{\rm{hie}}_{ij}\}_{j=0}^{N_v-1}$.
The hierarchical query of $j$-th point of $i$-th map element is formulated as:
\begin{equation}
\begin{gathered}
    q^{\rm{hie}}_{ij} =  q^{\rm{ins}}_i + q^{\rm{pt}}_j.
\end{gathered}
\end{equation}

\boldparagraph{Self-Attention Variants.}
MapTR~\cite{maptr} adopts vanilla self-attention to make hierarchical queries exchange information with each other (both inter-instance and intra-instance), whose computation complexity is $O((N\times N_{v})^2)$ ($N$ and $N_v$ are respectively the numbers of instance queries and point queries).
With the query number increasing, the computation cost and memory consumption dramatically increase.

In MapTRv2, to reduce the budget of  computation and memory, we adopt decoupled self-attention, \ie, respectively performing attention  along the inter-ins. dimension and intra-ins. dimension, as shown in Fig.~\ref{fig:framework}.
The decoupled self-attention greatly reduces the memory consumption and computation complexity (from $O((N\times N_{v})^2)$ to $O(N^2+N_{v}^2)$),
and results in higher performance than vanilla self-attention.

Another variant is only performing inter-ins. self-attention.
With only inter-ins. self-attention, MapTRv2 also achieves comparable performance (see Sec.~\ref{sec:ablation}).

\boldparagraph{Cross-Attention Variants.}
Cross-attention in the decoder is designed to make map queries interact with input features. 
We study on three kinds of  cross-attention: BEV-based,  PV-based  and hybrid cross-attention.

For BEV-based cross-attention,  we adopt Deformable Attention~\cite{deformdetr,bevformer} to make hierarchical queries  interact with BEV features.
For 2D map construction, each query $q^{\rm{hie}}_{ij}$ predicts the 2-dimension normalized BEV coordinate $(x_{ij},y_{ij})$ of the reference point $p_{ij}$. 
For 3D map construction, each query $q^{\rm{hie}}_{ij}$ predicts the 3-dimension normalized 3D coordinate $(x_{ij},y_{ij},z_{ij})$ of the reference point $p_{ij}$. 
We then sample BEV features around the reference points and update queries.

Map elements are usually with irregular shapes and require long-range context. Each map element corresponds to a set of reference points $\{p_{ij}\}_{j=0}^{N_v-1}$ with flexible and dynamic distribution.
The reference points $\{p_{ij}\}_{j=0}^{N_v-1}$ can adapt to the arbitrary shape of map element and capture informative context for map element learning.

For PV-based cross-attention, 
we  project the reference points $p_{ij}$ to the  PV images and then sample features around the projected reference points.  Dense BEV features are depreciated.

Hybrid cross-attention is  a combination of the above two  cross-attention manners.  Ablation experiments about these cross-attention variants are presented in  Sec.~\ref{sec:ablation}.

\boldparagraph{Prediction Head.}
The prediction head is simple, consisting of  a classification branch and  a point regression branch.
The classification branch predicts instance class score. The point regression branch predicts the positions of the point sets $\hat{V}$. For each map element, it outputs a $2N_v$- or $3N_v$-dimension vector, which represents normalized 2d or 3d coordinates  of the $N_v$ points.

\section{Training}
\label{sec:one2one}
\subsection{Hierarchical Bipartite Matching}
\label{sec:one2one_matching}
MapTRv2 parallelly infers a fixed-size set of $N$ map elements in a single pass, following the end-to-end paradigm of query-based object detection and segmentation paradigm~\cite{detr,yolos,queryinst}. $N$ is set to be larger than the typical number of map elements in a scene. Let's denote the set of $N$ predicted map elements by $\hat{Y} = \{\hat{y}_i \}_{i=0}^{N-1}$. 
The set of ground-truth (GT) map elements is padded with $\varnothing$ (no object) to form a set with size $N$, denoted by $Y=\{y_i\}_{i=0}^{N-1}$.
$y_i=(c_i, V_i, \Gamma_i)$, where $c_i$, $V_i$ and $\Gamma_i$ are respectively the target class label, point set and permutation group of GT map element $y_i$. $\hat{y}_i=(\hat{p}_i, \hat{V}_i)$, where $\hat{p}_i$ and $\hat{V}_i$ are respectively the predicted classification score and predicted point set.
To achieve structured map element modeling and learning,
MapTRv2 introduces hierarchical bipartite matching as shown in Fig.~\ref{fig:one2one}, \ie,  performing instance-level matching and point-level matching in order.

\boldparagraph{Instance-Level Matching.\label{sec:ins-lvl-matching}}
First, we need to find an optimal instance-level label assignment $\hat{\pi}$ between predicted map elements $\{\hat{y}_i \}$ and GT map elements $\{y_i \}$.  $\hat{\pi}$ is a permutation of $N$ elements ($\hat{\pi} \in \Pi_N$) with the lowest instance-level matching cost:
\begin{equation}
\begin{gathered}
\hat{\pi} = \argmin_{\pi \in \Pi_N} \sum_{i=0}^{N-1} \mathcal{L}_{\rm ins\_match} (\hat{y}_{\pi(i)}, y_i). \label{eq:argmin_inst_loss}
\end{gathered}
\end{equation}

$\mathcal{L}_{\rm ins\_match}(\hat{y}_{\pi(i)}, y_i)$ is a pair-wise matching cost between prediction $\hat{y}_{\pi(i)}$ and GT $y_i$, which  considers both the class label of map element and the position of point set:
\begin{equation}
\begin{gathered}
\mathcal{L}_{\rm ins\_match}(\hat{y}_{{\pi}(i)},  y_i ) = 
\mathcal{L}_{\rm{Focal}}(\hat{p}_{{\pi}(i)}, c_i) + \\
\hspace{10em}\mathcal{L}_{\rm position}(\hat{V}_{\pi(i)}, V_i).
\end{gathered}
\end{equation}
$\mathcal{L}_{\rm{Focal}}(\hat{p}_{{\pi}(i)}, c_i)$ is the class matching cost term, defined as the Focal Loss~\cite{focal} between predicted classification score $\hat{p}_{{\pi}(i)}$ and target class label $c_i$.
$\mathcal{L}_{\rm position}(\hat{V}_{\pi(i)}, V_i)$ is the position matching cost term, which reflects the position correlation between the predicted point set $\hat{V}_{\pi(i)}$ and the GT point set $V_i$. 
Hungarian algorithm is utilized to find the optimal instance-level assignment $\hat{\pi}$ following DETR~\cite{detr}.

\boldparagraph{Point-Level Matching.}
After instance-level matching,  each predicted map element $\hat{y}_{\hat{\pi}(i)}$  is assigned with a GT map element $y_i$.
Then for each predicted instance assigned with positive labels ($c_i\neq\varnothing$), we perform point-level matching
to find an optimal point-to-point assignment $\hat{\gamma} \in \Gamma$ between predicted point set $\hat{V}_{\hat{\pi}(i)}$ and GT point set $V_{i}$. $\hat{\gamma}$ is selected among the predefined permutation group ${\Gamma}$ and with the lowest point-level matching cost:
\begin{equation}
\begin{gathered}
\hat{\gamma} = \argmin_{\gamma \in {\Gamma}} \sum_{j=0}^{N_v-1} 
D_{\rm{Mht}}(\hat{v}_{j}, v_{ \gamma(j)}). \label{eq:argmin_pts_loss}
\end{gathered}
\end{equation}
$D_{\rm{Mht}}(\hat{v}_{j}, v_{ \gamma(j)})$ is the Manhattan distance between the $j$-th point of the predicted point set $\hat{V}$ and the $\gamma(j)$-th point of the GT point set $V$. 

\subsection{One-to-One Set Prediction Loss}
\label{sec:one2one_loss}
MapTRv2 is trained based on the optimal instance-level and point-level assignment ($\hat{\pi}$ and $\{\hat{\gamma_i}\}$).
The basic loss function is composed of three parts, classification loss, point-to-point loss and edge direction loss:
\begin{equation}
\begin{aligned}
\mathcal{L}_{\rm one2one} = &\mathcal{L}_{\rm Hungarian}(\hat{Y},Y)  \\
 = &\lambda_c \mathcal{L}_{\rm{cls}} + \lambda_p \mathcal{L}_{\rm{p2p}} + \lambda_d \mathcal{L}_{\rm{dir}},
\end{aligned}    
\end{equation}
where $\lambda_c$, $\lambda_p$ and $\lambda_d$ are the weights for balancing different loss terms.

\boldparagraph{Classification Loss.}
With  the  instance-level  optimal matching result $\hat{\pi}$,
each predicted map element is assigned with a class label \oldnew{(or 'no object' $\varnothing$)}{}. The classification loss is a Focal Loss term formulated as:
\begin{equation}
\begin{gathered}
\mathcal{L}_{\rm{cls}} = \sum_{i=0}^{N-1} \mathcal{L}_{\rm{Focal}}(\hat{p}_{\hat{\pi}(i)}, c_i).
\end{gathered}
\end{equation}

\boldparagraph{Point-to-Point Loss.}
Point-to-point loss supervises the position of each predicted point.
For each GT instance with index $i$,
according to  the  point-level optimal matching result $\hat{\gamma}_i$,
each predicted point $\hat{v}_{\hat{\pi}(i),j}$ is assigned with a GT point $v_{i, \hat{\gamma}_i(j)}$. 
The point-to-point loss is defined  as the Manhattan distance computed between each assigned point pair:

\begin{equation}
\begin{gathered}
    \mathcal{L}_{\rm{p2p}} = \sum_{i=0}^{N-1}   \mathbbm{1}_{\{c_i\neq \varnothing\}}    \sum_{j=0}^{N_v-1} D_{\rm {Mht}}(\hat{v}_{\hat{\pi}(i),j},  v_{i, \hat{\gamma}_i(j)}).
\end{gathered}    
\end{equation}

\boldparagraph{Edge Direction Loss.}
Point-to-point loss only supervises the node point of polyline and polygon, not considering the edge (the connecting line between adjacent points).
For accurately representing map elements, the direction of the edge is
important.
Thus, we further design edge direction loss to
supervise the geometrical shape in the higher edge level.
Specifically, we consider the  cosine similarity of the paired predicted edge $\boldsymbol{\hat{e}_{\hat{\pi}(i),j}}$ and GT edge $\boldsymbol{e_{i, \hat{\gamma}_i(j)}}$:
  
\begin{equation}
\begin{gathered}
    \mathcal{L}_{\rm{dir}} = - \sum_{i=0}^{N-1}   \mathbbm{1}_{\{c_i\neq \varnothing\}}  
    \sum_{j=0}^{N_v-1}  {\rm cos\_sim}(\boldsymbol{\hat{e}_{\hat{\pi}(i),j}}, \boldsymbol{e_{i, \hat{\gamma}_i(j)}}),\\
    \boldsymbol{\hat{e}_{\hat{\pi}(i),j}}=\hat{v}_{\hat{\pi}(i),j}  - \hat{v}_{\hat{\pi}(i),(j+1) \rm{mod} N_v},\\
    \boldsymbol{e_{i, \hat{\gamma}_i(j)}}=v_{i, \hat{\gamma}_i(j)} - v_{i, \hat{\gamma}_i(j+1) \rm{mod} N_v}.
\end{gathered}    
\end{equation}

\subsection{Auxiliary One-to-Many Set Prediction Loss}

\begin{figure}[h]
    \centering
    \includegraphics[width=0.98\linewidth]{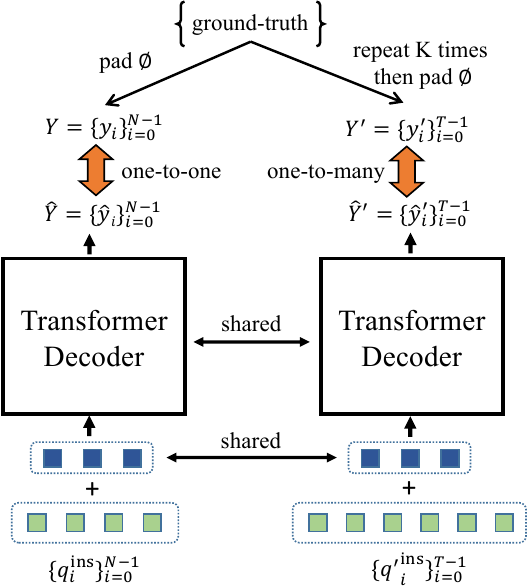}
    \caption{Auxiliary one-to-many matching branch. We introduce extra $T$ instance-level queries during training to provide auxiliary supervision. The point-level queries and Transformer decoder layers are shared by the two groups of instance queries. We perform one-to-one matching on the first group of instance queries, as demonstrated in Sec.~\ref{sec:one2one}, and perform one-to-many matching on the second group of instance queries by repeating the ground-truth map elements for $K$ times. The detailed matching procedure is the same as hierarchical bipartite matching.}
    \label{fig:one2many}
\end{figure}
To speed up the convergence, we add an auxiliary one-to-many matching branch during training, inspired by \cite{hybrid-detr}.
As shown in Fig.~\ref{fig:one2many}, 
the one-to-many matching branch shares  the same point queries and Transformer decoder with one-to-one matching branch,
but own an extra group of instance queries $\{q_i^{'\rm ins}\}_{i=0}^{T-1}$ ($T$ is the number).
This branch predict map elements $\hat{Y}^{'}=\{ \hat{y}^{'}_i \}_{i=0}^{T-1}$.

We repeat the ground-truth map elements  for $K$ times and pad them with $\varnothing$ to form a set with size $T$, denoted by $Y^{'} = \{ y^{'}_i\}_{i=0}^{T-1}$.
Then we perform the same hierarchical bipartite matching between $\widetilde{Y}$ and $\overline{Y}$,
and compute the auxiliary one-to-many set prediction loss:
\begin{equation}
\begin{aligned}
    \mathcal{L}_{\rm one2many} = &\mathcal{L}_{\rm Hungarian}(\hat{Y}^{'},Y^{'}) 
\end{aligned}    
\end{equation}

In the one-to-many matching branch, one GT element is assigned to $K$ predicted  elements.  With  the ratio of positive samples increasing, the map decoder converges faster.

\subsection{Auxiliary Dense Prediction Loss}
To further leverage semantic and geometric information,
we introduce three auxiliary dense prediction losses:
\begin{equation}
    \begin{aligned}
    \mathcal{L}_{\rm dense} = &\alpha_d \mathcal{L}_{\rm{depth}}  + \alpha_b \mathcal{L}_{\rm{BEVSeg}} + \alpha_p \mathcal{L}_{\rm{PVSeg}}.
\end{aligned}    
\end{equation}

\boldparagraph{Depth Prediction Loss.} Following BEVDepth~\cite{bevdepth}, we utilize the LiDAR point clouds to render the GT depth maps $\{D_1, \ldots, D_M\} $ of each perspective view. And we add a simple depth prediction head $\phi_{\rm depth}$ on the PV feature map $\mathcal{F}=\{F_1, \ldots, F_M\}$. The depth prediction loss is defined as the cross-entropy loss between the predicted depth map and the rendered GT depth map:
\begin{equation}
\begin{aligned}
    \mathcal{L}_{\rm depth} = \sum_{m=1}^{M}\mathcal{L}_{\rm CE}(\phi_{\rm depth}(F_m),D_m).
\end{aligned}    
\end{equation}

\boldparagraph{BEV Segmentation Loss.} 
Inspired by BeMapNet~\cite{qiao2023end}, we add an auxiliary BEV segmentation  head $\phi_{\rm BEVSeg}$ base on the BEV feature map $F_{BEV}$.  We rasterize the  map GT on the BEV canvas to get BEV foreground mask $M_{BEV}$. 
The BEV segmentation loss is defined as the cross entropy loss between the predicted BEV segmentation map and the binary GT map mask:
\begin{equation}
    \begin{aligned}
        \mathcal{L}_{\rm BEVSeg} = \mathcal{L}_{\rm CE}(\phi_{\rm BEVSeg}(F_{BEV}),M_{BEV}).
    \end{aligned}    
    \end{equation}
\boldparagraph{PV Segmentation Loss.} To fully exploit the dense supervision, we render  map GT on the perspective views with intrinsics and extrinsics of cameras and get perspective foreground masks $\{M^1_{PV}, \ldots, M^1_{PV}\}$.
And we add a auxiliary PV segmentation head $\phi_{\rm PVSeg}$ on the PV feature map $\mathcal{F}=\{F_1, \ldots, F_M\}$. The PV segmentation loss $\mathcal{L}_{\rm PVSeg}$ is defined as:
\begin{equation}
    \begin{aligned}
        \mathcal{L}_{\rm PVSeg} = \sum_{m=1}^{M}\mathcal{L}_{\rm CE}(\phi_{\rm PVSeg}(F_m),M^m_{PV}).
    \end{aligned}    
    \end{equation}
\subsection{Overall Loss}
The overall loss is defined as the weighted sum of the above losses:
\begin{equation}
    \begin{aligned}
    \mathcal{L} =  \beta_o \mathcal{L}_{\rm{one2one}} + \beta_m \mathcal{L}_{\rm{one2many}} +  \beta_d \mathcal{L}_{\rm{dense}}.
\end{aligned}    
\end{equation}

\begin{table*}[h]
    \caption{Comparisons with state-of-the-art methods~\cite{vectormapnet,hdmapnet} on nuScenes \texttt{val} set.  ``C'' and ``L'' respectively denotes camera and LiDAR. ``ft'' means fine-tuning, a training tirck illustrated in VectorMapNet~\cite{vectormapnet}. ``Effi-B0'', ``R18'', ``R50'', ``SwinT'', ``V2-99'', ``PP'' and ``Sec'' respectively correspond to EfficientNet-B0~\cite{efficientnet}, ResNet18~\cite{resnet}, ResNet50~\cite{resnet}, SwinTransformer-Tiny~\cite{swin} VoVNetV2-99~\cite{vovnet}, PointPillars~\cite{pointpillars} and LiDAR backbone used in SECOND~\cite{second}.
    Effi-B0, Effi-B4, SwinT, R50 and R18 are initialized with ImageNet~\cite{imagenet} pre-trained weights. V2-99 is initialized from DD3D~\cite{pseudolidar} checkpoint. Other LiDAR backbones are trained from scratch.
    The APs of other methods are taken from the papers~\cite{vectormapnet}. FPSs are measured on the same machine with RTX 3090. ``-'' means that the corresponding results are not available.}
    \centering
    \begin{tabular}{l|ccc|cccc|c}
    \toprule
    \multirow{2}{*}{Method} & \multirow{2}{*}{Modality} &  \multirow{2}{*}{Backbone} & \multirow{2}{*}{Epoch} & \multicolumn{4}{c|}{AP} & \multirow{2}{*}{FPS} \\
     &  &  &  & ped. & div. & bou. & mean & \\

    \midrule
    \multirow{3}{*}{HDMapNet} 
     & C & Effi-B0 & 30 & 14.4 & 21.7 & 33.0 & 23.0 & 0.9 \\
     & L & PP & 30 & 10.4 & 24.1 & 37.9 & 24.1& 1.1\\
     & C \& L & Effi-B0 \& PP & 30 & 16.3 & 29.6 & 46.7 & 31.0 & 0.5\\
    \midrule
    \multirow{3}{*}{VectorMapNet} 
     & C & R50& 110+ft & 42.5 & 51.4 & 44.1 & 46.0 &2.2\\
     & L & PP & 110 & 25.7 & 37.6 & 38.6 & 34.0 &-\\
     & C \& L & R50 \& PP & 110+ft & 48.2 & 60.1 & 53.0 & 53.7 &-\\
    \midrule
    InstaGraM & C & EffiNet-B4 & 30 & 33.8 & 47.2 & 44.0 & 41.7 & - \\
    \midrule
    \multirow{3}{*}{MapVR}  & C & R50 & 110 & 55.0 &61.8& 59.4& 58.8 & 15.1 \\
     & C & R50 & 24 & 47.7 &54.4 &51.4 &51.2& 15.1 \\
     & C \& L & R50 \& Sec & 24 & 60.4 & 62.7& 67.2 &63.5 & 6.0\\
    \midrule
    \multirow{2}{*}{BeMapNet} & C & R50 & 110 & 62.6 &66.7& 65.1& 64.8 & 7.0 \\
     & C & R50 & 30 & 57.7& 62.3& 59.4& 59.8 & 7.0 \\
    \midrule
    \multirow{3}{*}{PivotNet} & C & SwinT & 110 & 68.0 &62.6 &69.7 &66.8 & 8.3 \\
    & C & SwinT & 30 & 63.8 &58.7 &64.9 &62.5 & 8.3 \\ 
    & C & R50 & 30 & 53.8 &55.8& 59.6 &57.4 & 10.4 \\
    \midrule
    \multirow{5}{*}{MapTR} 
     & C & R18 & 110 & 39.6 & 49.9 & 48.2 & 45.9 &  \textbf{35.0}\\
     & C & R50 &110 & 56.2 & 59.8  & 60.1 &58.7 &15.1\\
     & C &R50 & \textbf{24} & 46.3 & 51.5 & 53.1 & 50.3& 15.1\\
     & L & Sec & \textbf{24} & 48.5 & 53.7 & 64.7 & 55.6& 8.0\\
     & C \& L & R50 \& Sec & \textbf{24} & 55.9 & 62.3 & 69.3 & 62.5& 6.0\\
    \midrule
    \multirow{7}{*}{MapTRv2}
     & C & R18 & 110 & 46.9 & 55.1 & 54.9 & 52.3 &  33.7\\
     & C & R50 & 110 & 68.1 & 68.3 & 69.7 & 68.7 &  14.1\\
     & C & V2-99 & 110 & \textbf{71.4} & \textbf{73.7} & 75.0 & 73.4 &  9.9\\
     & C & R50 & \textbf{24} & 59.8 & 62.4 & 62.4 & 61.5 &  14.1\\
     & C & V2-99 & \textbf{24} & 63.6 & 67.1 & 69.2 & 66.6 &  9.9\\
     & L & Sec & \textbf{24} & 56.6 & 58.1 & 69.8 & 61.5 &  7.6\\
     & C \& L & R50 \& Sec & \textbf{24} & 65.6 & 66.5 & 74.8 & 69.0 &  5.8\\
     & C \& L & V2-99 \& Sec & \textbf{24} & 70.4&73.2&\textbf{78.6}&\textbf{74.0} &  4.5\\

    \bottomrule
    \end{tabular}
    \label{tab:main-result}
    \end{table*}

    \begin{table*}[t!]
    \caption{Comparisons with state-of-the-art methods~\cite{vectormapnet,hdmapnet} on Argoverse2 \texttt{val} set.  Argoverse2 provides 3D vectorized map, which has extra height information compared to nuScenes dataset. $\rm{dim} = 2$ denotes that the height information of vectorized map is dropped. $\rm{dim} = 3$ denotes directly predicting 3D vectorized map. The results of VectorMapNet and HDMapNet are taken from VectorMapNet paper~\cite{vectormapnet}. FPSs are measured on the same machine with RTX 3090. ``-'' means that the corresponding results are not available.}
    \centering
    \begin{tabular}{l|ccc|cccc|c}
    \toprule
    \multirow{2}{*}{Method} & \multirow{2}{*}{Map dim.} &  \multirow{2}{*}{Backbone} & \multirow{2}{*}{Epoch} & \multicolumn{4}{c|}{AP} & \multirow{2}{*}{FPS} \\
        &  &  &  & ped. & div. & bou. & mean & \\

    \midrule
    HDMapNet
        & 2 & Effi-B0 & - & 13.1 & 5.7 & 37.6 & 18.8 & - \\
    \midrule
    \multirow{2}{*}{VectorMapNet} 
        & 2 & R50 & - & 38.3 & 36.1 & 39.2 & 37.9 & - \\
        & 3 & R50 & - & 36.5 & 35.0 & 36.2 & 35.8 & -\\
    \midrule
    MapVR & 2 & R50 & 6 & 54.6 & 60.0& 58.0& 57.5 & - \\
    \midrule
    \multirow{2}{*}{MapTRv2}
        & 2 & R50 & 6 & 62.9 & 72.1 & 67.1 & 67.4 &  12.1\\
        & 3 & R50 & 6 & 60.7 & 68.9 & 64.5 & 64.7 &  12.0\\

    \bottomrule
    \end{tabular}
    \label{tab:av2}
    \end{table*}
\section{Experiments}
In this section, we first introduce our experimental setup and implementation details in Sec.~\ref{sec:exp-setup}. Then we present the main results of our frameworks and compare with state-of-the-art methods in Sec.~\ref{sec:main-results}. And in Sec.~\ref{sec:ablation}, we conduct extensive ablation experiments and analyses to investigate the individual components of our framework. In Sec.~\ref{sec:main-results} and Sec.~\ref{sec:centerline}, we show that our framework is easily to be extended to 3D map construction and centerline learning.
\subsection{Experimental Setup}
\label{sec:exp-setup}
\boldparagraph{Datasets.} 
We mainly evaluate our method on the popular nuScenes~\cite{nuscenes} dataset following the standard setting of previous methods~\cite{hdmapnet,vectormapnet,maptr}. The nuScenes dataset contains  2D city-level global vectorized maps and 1000 scenes of roughly 20s duration each. Key samples are annotated at $2$Hz. Each sample has RGB images from $6$ cameras and covers $360^\circ$ horizontal FOV of the ego-vehicle. 

We further provide experiments on Argoverse2~\cite{av2} dataset, which contains 1000 logs. Each log provides 15s of 20Hz RGB images from 7 cameras and a log-level 3D vectorized map.

\boldparagraph{Metric.}
We follow the standard metric used in previous works~\cite{hdmapnet,vectormapnet,maptr}. The perception ranges are $[-15.0m, 15.0m]$ for the $X$-axis and $[-30.0m, 30.0m]$ for the $Y$-axis. And we adopt average precision (AP) to evaluate the map construction quality.
Chamfer distance $D_{Chamfer}$ is used to determine whether the prediction and GT are matched or not.
We calculate the $\rm{AP}_{\tau}$ under several $D_{Chamfer}$ thresholds ($\tau \in T, T=\{0.5, 1.0, 1.5\}$), and then average across all thresholds as the final AP metric:

\begin{equation}
\rm{AP} = \frac{1}{|T|} \sum_{\tau \in T} \rm{AP}_{\tau}.
\end{equation}

Following the previous methods ~\cite{hdmapnet,vectormapnet,maptr}, three kinds of map elements are chosen for fair evaluation~\cite{vectormapnet,maptr,hdmapnet} -- pedestrian crossing, lane divider, and road boundary.
Besides,  we also extend MapTRv2 to modeling and learning centerline (see Sec.~\ref{sec:centerline}) and provide additional evaluation.

\boldparagraph{Training Details.} ResNet50~\cite{resnet} is used as the image backbone network unless otherwise specified. The optimizer is AdamW with weight decay 0.01. The batch size is 32 (containing 6 view images) and all models are trained with 8 NVIDIA GeForce RTX 3090 GPUs. Default training schedule is 24 epochs and the initial learning rate is set to $6\times 10^{-4}$ with cosine decay. We extract ground-truth map elements in the perception range of ego-vehicle following~\cite{maptr,vectormapnet,hdmapnet}. The resolution of source nuScenes images is $1600 \times 900$. For the real-time version, we resize the source image with 0.2 ratio and the other models are trained with 0.5-ratio-resized images. 
For Argoverse2 dataset, the 7 camera images have different resolutions ($1550\times2048$ for front view and $2048\times1550$ for others). We first pad the 7 camera images into the same shape ($2048\times2048$), then resize the images with 0.3 ratio.
Color jitter is used by default in both nuScenes dataset and Argoverse2 dataset. The default number of instance queries, point queries and decoder layers is 50, 20 and 6, respectively. For PV-to-BEV transformation, we set the size of each BEV grid to 0.3m and utilize efficient BEVPoolv2~\cite{bevpoolv2} operation. Following~\cite{maptr}, $\lambda_c=2$, $\lambda_p=5$, $\lambda_d=0.005$. For dense prediction loss, we set $\alpha_d$, $\alpha_p$, $\alpha_b$ to 3, 2 and 1 respectively. For the overall loss, $\beta_o=1$, $\beta_m=1$, $\beta_d=1$. We also provide long training schedule (110 epochs) on nuScenes dataset without changing other hyper-parameters for fair comparison with previous methods~\cite{vectormapnet}.

\boldparagraph{Inference Details.} The inference stage is quite simple. Given surrounding images, we can directly predict 50 map elements associated with their scores. The scores indicate the confidence of predicted map element. The top-scoring predictions can be directly utilized without other post-processing. The inference time is measured on a single NVIDIA GeForce RTX 3090 GPU with batch size 1.

\subsection{Main results}
\label{sec:main-results}

\boldparagraph{Comparison with State-of-the-Art Methods.}
We train MapTRv2 under 24-epoch schedule and 110-epoch schedule on nuScenes dataset. As shown in Table~\ref{tab:main-result}, MapTRv2 outperforms all the state-of-the-art methods by a large margin in terms of convergence, accuracy and speed. Surprisingly, MapTRv2 based on ResNet-50 achieves 68.7 mAP, which 10 mAP higher than corresponding MapTR counterpart with similar speed  and 6 mAP higher than the previous best-performing multi-modality result while being 2x faster.
MapTRv2-VoVNet99 with only camera input  achieves 73.4 mAP, which is 19.7 mAP higher than multi-modality VectorMapNet while being 4x faster and 10.9 mAP higher than multi-modality MapTR as well as 3.9 FPS faster. When compared with the concurrent method PivotNet-SwinT~\cite{pivotnet}, MapTRv2-VoVNet99 outperforms by 6.6 mAP and 1.6 FPS.
As shown in Fig.~\ref{fig:tradeoff}, MapTRv2 achieves the best trade-off between accuracy and speed. MapTRv2-ResNet18 achieves 50+ mAP at the real-time speed.

\begin{figure}[h]
    \centering
    \includegraphics[width=0.98\linewidth]{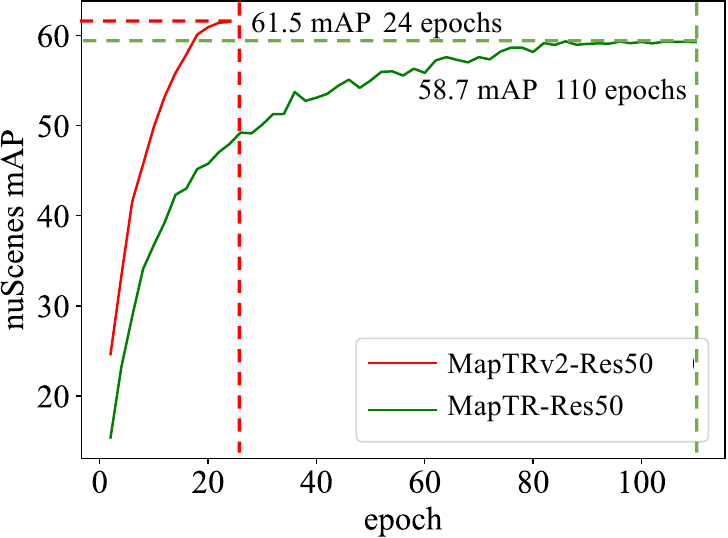}
    \caption{Convergence curves of MapTRv2 and MapTR with ResNet-50  backbone on nuScenes dataset. MapTRv2 significantly outperforms  MapTR with 4x shorter training schedule and 2.8 higher mAP.}
    \label{fig:convergence}
\end{figure}
Meanwhile, we also evaluate our framework on the Argoverse2 dataset. Argoverse2 provides 3D vectorized map, which has extra height information compared to nuScenes dataset.
As shown in Table~\ref{tab:av2},  In terms of 2D vectorized map construction, MapTRv2 achieves 67.4 mAP, which is 29.5 mAP higher than VectorMapNet, 48.6 mAP higher than HDMapNet and 9.9 mAP higher than MapVR. In terms of 3D vectorized map construction, MapTRv2 achieves 64.7 mAP, which is 29.9 mAP higher than VectorMapNet.
The experiments on Argoverse2 dataset demonstrates the superior generalization ability of MapTRv2 in terms of 3D vectorized map construction.

\begin{table*}[t!]
	\caption{Roadmap of MapTRv2 on nuScenes val set with only 24-epoch training schedule.}
    \centering
    \begin{tabular}{l | c c c|l}
        \toprule
        \multirow{2}{*}{Method}  & \multicolumn{3}{c|}{AP} & \multirow{2}{*}{mAP} \\
                    & ped. & div. & bou. &  \\ \midrule
        MapTR baseline         & 46.3   & 51.5  & 53.1 & 50.3 \\
        \ \  \small{+2D-to-BEV with LSS}     & 44.9    & 51.9  & 53.5  & 50.1 \color{ForestGreen}\small \textbf{(-0.2)}   \\
        \ \ \small{+auxiliary depth sup.}     & 49.6    & 56.6  & 59.4   & 55.2 \color{ForestGreen}\small \textbf{(+4.9)} \\
        \ \  \small{+auxiliary BEV sup.}      & 52.1    & 57.6  & 59.9   & 56.5 \color{ForestGreen}\small \textbf{(+6.2)}  \\
        \ \  \small{+auxiliary PV sup.}       & 53.2    & 58.1  & 60.0   & 57.1 \color{ForestGreen}\small \textbf{(+6.8)} \\
        \ \ \small{+decoupled self-attn.}      & 53.9    & 58.3  & 60.5   & 57.6 \color{ForestGreen}\small \textbf{(+7.3)}  \\
        \ \ \small{+one-to-many matching}     & 59.8   & 62.4  & 62.4   & 61.5 \color{ForestGreen}\small \textbf{(+11.1)}   \\
        \textcolor[RGB]{128,138,135}{\ \ \small{+BEV free (optional)}}          & \textcolor[RGB]{128,138,135}{42.5}    & \textcolor[RGB]{128,138,135}{52.0}  & \textcolor[RGB]{128,138,135}{53.4}   & \textcolor[RGB]{128,138,135}{49.5}  \small \textcolor[RGB]{128,138,135}{(-12.0)}  \\ \bottomrule
        \end{tabular}

	\label{tab:roadmap}
\end{table*}

\begin{table*}[t!]
    \vspace*{-0.5cm}
    \caption{Ablation on the permutation-equivalent shape modeling. Our proposed permutation-equivalent shape modeling outperforms the fixed-order shape modeling by 4.1 mAP and 7.1 AP in pedestrian-crossing especially.}
    \centering
    \begin{tabular}{l| c c c c}
            \toprule
            \multirow{2}{*}{Shape Modeling} & \multicolumn{4}{c}{AP} \\
                        & ped. & div. & bou. & mean   \\ \midrule
            Fixed-order $V^F$ w/ ambiguity & 52.7 & 59.1 & 60.3 & 57.4 \\
            Permutation-equivalent $(V,\Gamma)$ w/o ambiguity & \cellcolor{gray!30}59.8 & \cellcolor{gray!30}62.4 & \cellcolor{gray!30}62.4 & \cellcolor{gray!30}61.5 \\
            \bottomrule
            \end{tabular}
    \label{tab:shape_modeling}
\end{table*}

\boldparagraph{Roadmap of MapTRv2.}
In Table~\ref{tab:roadmap}, we show how we build MapTRv2 upon MapTR.  We first replace the default PV-to-BEV transformation method GKT in MapTR with LSS. Then we gradually add the auxiliary dense supervision, memory-efficient decoupled self-attention and auxiliary one-to-many matching. With all these components, MapTRv2-ResNet50 achieves 61.5 mAP, which is 11.1 mAP higher than MapTR-ResNet50 with similar speed. As shown in Fig.~\ref{fig:convergence}, MapTRv2 achieves superior convergence than MapTR (24 epochs, 61.5 mAP vs. 110 epochs, 58.7mAP).
We also provide a BEV-free variant of MapTRv2 by replacing the BEV-based cross-attention with PV-based cross-attention and removing PV-to-BEV transformation module. Though the performance of BEV-free MapTRv2 drops, this variant gets rid of costly dense BEV representation and is more lightweight.

\subsection{Ablation Study}
\label{sec:ablation}
In this section, we investigate the individual components of our MapTRv2 framework with extensive ablation experiments. We conduct all experiments on the popular nuScenes dataset unless otherwise specified. For efficiently conducting the ablation experiments, we use MapTRv2-ResNet50 with  24-epoch training schedule  as the baseline model unless otherwise specified.

\boldparagraph{Permutation Equivalent Modeling.}
In Table~\ref{tab:shape_modeling}, we provide ablation experiments to validate the effectiveness of the proposed permutation-equivalent modeling. Compared with vanilla modeling method which imposes a unique permutation to the point set, permutation-equivalent modeling solves the ambiguity of map element and brings an improvement of 4.1 mAP. For pedestrian crossing, the improvement even reaches 7.1 AP, proving the superiority in modeling polygon elements.

\boldparagraph{Self-Attention Variants.}
In this ablation study, we remove the hybrid matching to fully reveal the efficiency of self-attention variants. As shown in Table~\ref{tab:self-attn}, the instance self-attention  largely reduces the training memory (2265M memory reduction) while at cost of negligible accuracy drop (0.3 mAP drop). To recover the interaction in each instance, decoupled self-attention  adds one more self-attention on point-dim. It is  much more memory-efficient (1985M memory reduction) and achieves better accuracy (0.5 mAP higher) and similar speed compared with vanilla self-attention. To further reveal superiority of the decoupled self-attention, in Table~\ref{tab:query_num on self-attn},
we compare vanilla self-attention with decoupled self-attention in terms of memory consumption and the number of instance queries.
with the number of instance queries increasing, mAP and memory cost consistently increases.   Under the same memory limitation (24 GB of RTX 3090), decoupled self-attention requires much less memory and results in higher performance upper bound.

\begin{table}[t!]
    \caption{Ablation on the self-attention variants. The inter self-attention  demonstrates significant memory reduction while maintaining comparable accuracy.  We choose the decoupled self-attention as the default configuration.}
    \setlength{\tabcolsep}{2.5pt}
    \begin{tabular}{l| c c c c | r | c}
            \toprule
            \multirow{2}{*}{Self-Attn} & \multicolumn{4}{c|}{AP} &\multirow{2}{*}{GPU mem.} & \multirow{2}{*}{FPS} \\
                        & ped. & div. & bou. & mean & & \\ \midrule
            Vanilla & 53.2 & 58.1 & 60.0 & 57.1 & 10443 M& 14.7 \\
            Only inter-ins. & 53.1 & 57.2 & 60.2 & 56.8 & 8178 M& 14.9\\
            Decoupled  & 53.9 & 58.3 & 60.5 & 57.6 & 8458 M& 14.1 \\
            \bottomrule
            \end{tabular}
    \vspace*{-0.7cm}
    \label{tab:self-attn}
\end{table}

\begin{table}[t!]
    \caption{Comparisons between vanilla self-attention and decoupled self-attention in terms of memory consumption and number of instance queries. ``+'' denotes the results of vanilla self-attention, ``*'' denotes the results of decoupled self-attention. ``OOM'' denotes out of memory. ``-'' denotes the results are not available. }
    \setlength{\tabcolsep}{3pt}
    \begin{tabular}{l | ccccc}
            \toprule
            N &  50 & 75 & 100 & 125 & 150\\
            \midrule
            mAP$^{+}$ &  57.1 & 58.4 & 59.5 & - & -\\
            GPU memory$^{+}$ (M) & 10443 & 13720 & 18326 & OOM & OOM\\
            \midrule
            mAP$^{*}$ & 57.6 & 60.1 & 60.3 & 60.4 & 61.2\\
            GPU memory$^{*}$ (M) & 8458 & 9049 & 9670 & 10378 & 11157\\
            \bottomrule
            \end{tabular}

    \label{tab:query_num on self-attn}
\end{table}

\boldparagraph{Cross-Attention Variants.}
In Table~\ref{tab:cross-attn}, we ablate on  cross-attention on nuScenes dataset and Argoverse2 dataset. Height information of map is not available on nuScenes dataset, thus the projected reference points on PV features are not accurate enough.  When we  simply replace the BEV-based deformable cross-attention with the PV-based one,  the accuracy drops by 12.0 mAP. When we stack the BEV-based and PV-based cross-attention, the performance is still inferior to that of BEV-based one. 
But on Argoverse2 dataset,  height information of map is available and the projected reference points can be supervised to be accurate. So the performance gap between BEV-based and PV-based cross-attention is much smaller compared to that on nuScenes. And the hybrid cross-attention  improves the BEV-based accuracy by 0.9 mAP, demonstrating the complementarity of PV features and BEV features.
\begin{table}[t!]
    \caption{Ablation on the cross-attention variants on nuScenes and Argoverse2. ``nus. 2D'' denotes that nuScenes dataset only provides 2D vectorized maps, lacking in height information. ``av2. 3D'' denotes that Argoverse2 provides 3D vectorized maps. ``BEV'' denotes BEV-based cross-attention. ``PV'' denotes PV-based cross-attention. ``BEV + PV'' denotes the hybrid cross-attention. We choose the BEV-based cross-attention as the default configuration. }
    \setlength{\tabcolsep}{4pt}
    \begin{tabular}{l|c| c c c c | c }
            \toprule
            \multirow{2}{*}{Dataset} &\multirow{2}{*}{Cross-Attn} & \multicolumn{4}{c|}{ AP}  & \multirow{2}{*}{FPS} \\
                      &  & ped. & div. & bou. & mean &  \\ \midrule
            \multirow{3}{*}{nus. 2D}
            & BEV &\cellcolor{gray!30}59.8 & \cellcolor{gray!30}62.4 & \cellcolor{gray!30}62.4 & \cellcolor{gray!30}61.5& \cellcolor{gray!30}14.1 \\
            & PV & 42.5 & 52.0 & 53.4 & 49.5& 14.4\\
            & BEV + PV  & 58.9 & 60.6 & 62.0 & 60.5 & 11.5 \\
            \midrule
            \multirow{3}{*}{av2. 3D}
            & BEV &\cellcolor{gray!30}60.7 & \cellcolor{gray!30}68.9 & \cellcolor{gray!30}64.5 & \cellcolor{gray!30}64.7& \cellcolor{gray!30}12.0 \\
            & PV & 53.1 & 63.4 & 60.8 & 59.1& 12.7\\
            & BEV + PV  & 61.4 & 69.9 & 65.6 & 65.6 & 10.0 \\
            \bottomrule
            \end{tabular}

            \vspace*{-0.2cm}
    \label{tab:cross-attn}
\end{table}

\boldparagraph{Dense Prediction Loss.}
In Table~\ref{tab:aux-sup}, we ablate on the effectiveness of dense supervision. The results show that depth supervision, PV segmentation and BEV segmentation provide complementary supervision. By adding them together, the final performance is improved by 4.9 mAP.
\begin{table}[t!]
    \caption{Ablation on the auxiliary dense prediction losses. ``Depth'' denotes depth prediction loss. ``Seg$^{\rm PV}$'' denotes PV segmentation loss. ``Seg$^{\rm BEV}$'' denotes BEV segmentation loss. Each of them provides complementary supervision.}
    \setlength{\tabcolsep}{3.5pt}
    \begin{tabular}{c c c | c c c c | c }
            \toprule
            \multirow{2}{*}{Depth} & \multirow{2}{*}{Seg$^{\text{PV}}$} & \multirow{2}{*}{Seg$^{\text{BEV}}$} & \multicolumn{4}{c|}{AP} & \multirow{2}{*}{FPS} \\
                  &  &   & ped. & div. & bou. & mean & \\ \midrule
            &  &  &  53.4& 58.8 & 57.5 & 56.6 & 14.1 \\
            \checkmark&  &  & 57.3 & 60.5 & 61.7 & 59.8 & 14.1 \\
            \checkmark & \checkmark & &  58.8 & 61.3 & 61.4 & 60.5 & 14.1\\
            \checkmark &  & \checkmark &  60.1 & 61.3 & 61.6 & 61.0 & 14.1\\
             & \checkmark & \checkmark & 57.5 & 60.5 & 59.7 & 59.2 & 14.1\\
            \cellcolor{gray!30}\checkmark & \cellcolor{gray!30}\checkmark & \cellcolor{gray!30}\checkmark & \cellcolor{gray!30}59.8 & \cellcolor{gray!30}62.4 & \cellcolor{gray!30}62.4 &\cellcolor{gray!30} 61.5 & \cellcolor{gray!30}14.1\\
            \bottomrule
            \end{tabular}
    
    \label{tab:aux-sup}
\end{table}

\begin{table*}[htbp!]
    \caption{Ablation on $K$ in hybrid matching. We set $T = 50 \times K$.}
    \centering
    \begin{tabular}{l | c c c c c c c}
    \toprule
    $K$ & 0 & 1 & 2 & 3 & 4 & 5 & \cellcolor{gray!30}6\\
    \midrule
    mAP& 57.6 & 58.7 & 61.0 & 60.9 & 60.9 & 60.9 & \cellcolor{gray!30}61.5\\
    GPU mem.(M) & 8458 & 9670 & 11157 & 12875 & 14814 & 17004 & \cellcolor{gray!30}19426\\
    FPS & 14.1 & 14.1 & 14.1 & 14.1 & 14.1 & 14.1 & \cellcolor{gray!30}14.1\\
    \bottomrule
    \end{tabular}
    \label{tab:hymatching-K}
\end{table*}
\begin{table*}[htbp!]
    \caption{Ablation on $T$ in hybrid matching. We set $K=6$.}
    \centering
    \begin{tabular}{l|ccccc }
    \toprule
    $T$ & 100 & 150 & 200 & 250 & \cellcolor{gray!30}300 \\
    \midrule
    mAP& 58.0 & 59.6 & 60.3 & 61.4 & \cellcolor{gray!30}61.5 \\
    GPU mem.(M) & 11157 & 12875 & 14814 & 17004 & \cellcolor{gray!30}19426\\
    FPS & 14.1 & 14.1 & 14.1 & 14.1 & \cellcolor{gray!30}14.1\\
    \bottomrule
    \end{tabular}
    \label{tab:hymatching-T}
\end{table*}
\boldparagraph{One-to-Many Loss.}
We ablate on the hyper-parameter  $K$ in Table~\ref{tab:hymatching-K}. With $K$ increasing, the training memory cost correspondingly increases.
Due to the limited GPU memory, we only increase $K$ up to 6, which brings an improvement of 3.9 mAP. And in Table~\ref{tab:hymatching-T}, we fix $K=6$ and ablate on $T$. The results show consistent improvement from $T=100$ to $T=300$. We further ablate on the hyper-parameter $\beta_m$ of auxiliary one-to-many loss in Table~\ref{tab:beta_m}.
Finally, we choose $T=300$, $K=6$, $\lambda=1$ as the default setting of MapTRv2.
\begin{table}[htbp!]
    \caption{Ablation on $\beta_m$. }
    \label{tab:beta_m}
    \begin{tabular}{@{}l | cccccc @{}}
    \toprule
    $\beta_m$ & 0.1 & 0.2 & 0.5 & \cellcolor{gray!30}1.0 & 2.0 & 5.0 \\
    \midrule
    mAP& 58.5 & 59.3 & 60.6 & \cellcolor{gray!30}61.5 & 59.9  & 60.4 \\
    \bottomrule
    \end{tabular}
\end{table}

\boldparagraph{Input Modality.}
As shown in Table~\ref{tab:main-result},  our framework is compatible with other vehicle-mounted sensors like LiDAR. Multi-modality MapTRv2 achieves superior performance (7.5 mAP higher) than single-modality MapTRv2 under 24-epoch training schedule, demonstrating the complementarity of different sensor inputs. And we reduce the gap between camera-only and LiDAR-only from 5.3 mAP in MapTR to 0.0 mAP in MapTRv2.

\boldparagraph{The Number of Points.}
In Table~\ref{tab:point-num}, we ablate on the point number of each element. The results show that 20 points are enough to achieve the best performance. Further increasing the point number increases the optimization difficulty and leads to performance drop. We choose 20 points as the default configuration.
\begin{table}
    \caption{Ablation on the number of point queries. }
    \label{tab:point-num}
    \begin{tabular}{l| c c c c | c | c}
    \toprule
    \multirow{2}{*}{Point} & \multicolumn{4}{c|}{AP} &\multirow{2}{*}{Param.} & \multirow{2}{*}{FPS} \\
                & ped. & div. & bou. & mean & & \\ \midrule
    10 & 55.3 & 64.1 & 60.5 & 60.0 & 39.0 & 14.1 \\
    \cellcolor{gray!30}20 & \cellcolor{gray!30}59.8 & \cellcolor{gray!30}62.4 &\cellcolor{gray!30}62.4 &\cellcolor{gray!30}61.5 & \cellcolor{gray!30}39.0 & \cellcolor{gray!30}14.1\\
    30  & 56.7 & 58.7 & 61.8 & 59.0 & 39.0 & 14.0 \\
    40  & 55.1 & 54.8 & 61.0 & 56.9 & 39.0 & 14.2 \\
    \bottomrule
    \end{tabular}
\end{table}

\boldparagraph{The Number of Transformer Decoder Layers.} Map decoder layers iteratively update the predicted map. Table~\ref{tab:layer-num} shows the ablation about the number of decoder layers. Without the iterative architecture, the result is 39.5 mAP. Increasing to 2 layers brings a gain of 13.1 mAP. The performance is saturated at 6 layers. We choose 6 layers as the default configuration.
\begin{table}[t!]
    \caption{Ablation on the number of Transformer decoder layers. }
    \label{tab:layer-num}
    \setlength{\tabcolsep}{5.5pt}
    \begin{tabular}{l| c c c c | c | c}
    \toprule
    \multirow{2}{*}{Layer} & \multicolumn{4}{c|}{AP} &\multirow{2}{*}{Param.} & \multirow{2}{*}{FPS} \\
                & ped. & div. & bou. & mean & & \\ \midrule			
    1 & 32.0 & 42.8& 43.7 & 39.5& 32.9 & 17.9 \\
    2 & 46.0 & 54.7 & 57.0 & 52.6 & 34.1 & 16.9\\
    3  & 53.2 & 57.6 & 59.7 & 56.8 & 35.3 & 16.0 \\
    \cellcolor{gray!30}6  & \cellcolor{gray!30}59.8 &\cellcolor{gray!30} 62.4 & \cellcolor{gray!30}62.4 & \cellcolor{gray!30}61.5 & \cellcolor{gray!30}39.0 & \cellcolor{gray!30}14.1 \\
    8  & 60.2 & 62.3 & 61.4 & 61.3 & 41.4 & 13.1 \\
    \bottomrule
    \end{tabular}
\end{table}

\begin{figure*}[t!]
    \centering
    \includegraphics[width=0.9\linewidth]{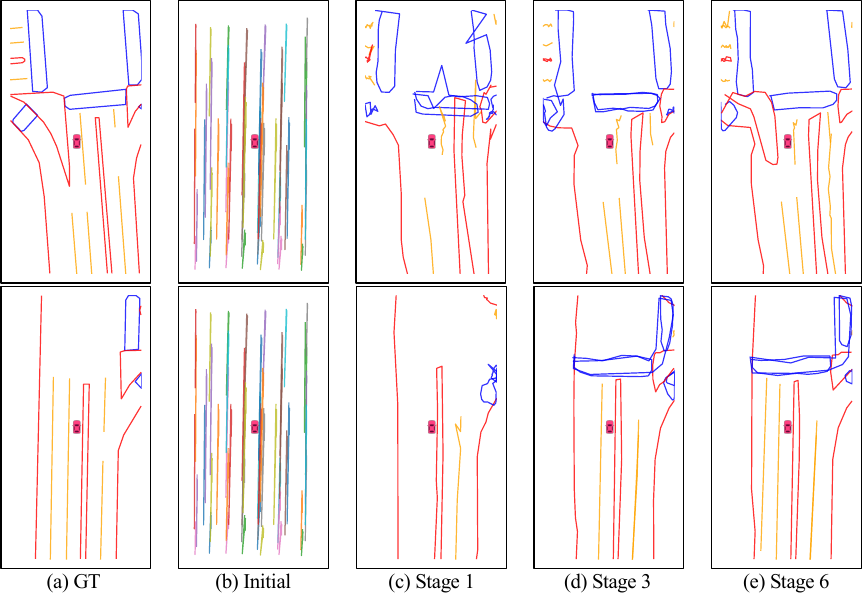}
    \caption{Iterative refinement. Column (a) is the ground-truth vectorized map. Column (b) is the map predicted by initial queries, in which different colors indicate different instances. Column (c) and (d) are the intermediate results after 1 and 3 Transformer decode layers, respectively. Column (e) is the final results. For column (c), (d) and (e), we only visualize those predicted map elements whose classification scores are above 0.3. The initial queries serve as priors of map. The cascade Transformer decoder layers iteratively refine positions and remove duplicate ones.}
    \label{fig:iterative}
\end{figure*}

\boldparagraph{Iterative Refinement.} In Fig.~\ref{fig:iterative}, we show the procedure of iterative refinement of a converged MapTRv2 baseline model by visualizing the predicted map elements at different decoder layers. 
The initial queries seem like evenly distributed polylines with  different lengths (Fig.~\ref{fig:iterative} (b)), which serve as the priors of map.
The cascaded decoder layers gradually refine the predictions and remove duplicate ones, as shown in Fig.~\ref{fig:iterative} (c), (d) and (e).

\boldparagraph{Detailed Runtime.}
In Table~\ref{tab:runtime}, we provide the detailed runtime of each component in MapTRv2-ResNet50 with only camera input. The results show that the main inference cost lies in the backbone network (59.8\%) while the other modules are quite efficient.

\begin{table}[t!]
    \caption{Detailed runtime of each component in MapTRv2-ResNet50 on one RTX 3090 GPU.}
    \label{tab:runtime}
    \begin{tabular}{l| c | c}
    \toprule
    Component & Runtime (ms) & Proportion \\
    \midrule
    Backbone & 42.4 & 59.8\% \\
    PV2BEV module & 11.6 & 16.4\% \\
    Map decoder & 16.9 & 23.8\% \\
    \midrule
    Total & 70.9 & 100\% \\
    \bottomrule
    \end{tabular}

\end{table}

\begin{table}[t!]
    \caption{Robustness to the translation deviation of camera. $\sigma_1$ is the standard deviation of  $\Delta_x, \Delta_y, \Delta_z$.}
    \label{tab:robustness-translation}
    \begin{tabular}{l | c c c c c}
    \toprule
    $\sigma_{1}(m)$ & 0 & 0.05 & 0.1 & 0.5 & 1.0 \\
    \midrule
    mAP& 61.5 & 60.8 & 58.0 & 50.6 & 32.3 \\
    \bottomrule
    \end{tabular}
\vspace*{-0.2cm}
\end{table}

\begin{table}[t!]
    \caption{Robustness to the rotation deviation of camera. $\sigma_2$ is the standard deviation of $\theta_x, \theta_y, \theta_z$.}
    \label{tab:robustness-rotation}
    \begin{tabular}{l | c c c c c}
    \toprule
    $\sigma_{2}(rad)$ & 0 & 0.005 & 0.01 & 0.02 & 0.05 \\
    \midrule
    mAP& 61.5 & 61.2 & 59.8 & 35.6 & 18.8 \\
    \bottomrule
    \end{tabular}
    \vspace*{-0.2cm}
\end{table}

\begin{table*}[htbp!]
    \caption{Performance with centerline learning. The mAP of centerline is as high as those of other elements. The results show that MapTRv2 is highly extensible and compatible to different map elements.}
    \label{tab:centerline}
    \centering
    \begin{tabular}{l| c| c| c c c c c }
    \toprule
    \multirow{2}{*}{Dataset} & \multirow{2}{*}{Dim} & \multirow{2}{*}{Epoch}& \multicolumn{5}{c}{AP}  \\
            & &    & ped. & div. & bou. & cen. & mean\\ \midrule
    nuScenes & 2 & 24 & 50.1 & 53.9 &58.8 &53.1&54.0\\
    Argoverse2  & 2 & 6 & 55.2 & 67.2 & 64.8 & 63.2 & 62.6 \\
    Argoverse2  & 3 & 6 & 53.5 & 66.9 & 63.6 & 61.5 &61.4\\
    \bottomrule
    \end{tabular}
    \vspace*{-0.2cm}
\end{table*}

\boldparagraph{Robustness to the Camera Deviation.} 
In real applications, the camera intrinsics are usually accurate and change little, but the camera extrinsics may be inaccurate due to the shift of camera position, calibration error, \etc. To validate the robustness, we traverse the validation sets and randomly generate noise for each sample. We respectively add translation and rotation deviation of different degrees.
Note that we add noise to all cameras and all coordinates. 
And the noise is subject to normal distribution. There exists extremely large deviation in some samples, which affect the performance a lot. As illustrated in Table~\ref{tab:robustness-translation} and Table~\ref{tab:robustness-rotation}, when the standard deviation of  $\Delta_x, \Delta_y, \Delta_z$ is  $0.1m$ or the standard deviation of $\theta_x, \theta_y, \theta_z$ is  $0.01rad$, MapTRv2 still keeps comparable performance.

\subsection{Extention: Centerline}
\label{sec:centerline}
Centerline can be seen as an special type of  map elements, which provide  directional information, indicate the traffic flow and play an important role in downstream planners as demonstrated in~\cite{nuplan_garage}. Following the path-wise modeling proposed in LaneGAP~\cite{lanegap}, we include centerline learning in MapTRv2.
As shown in Table~\ref{tab:centerline}, with centerline included, MapTRv2 achieves 54.0 mAP on nuScenes dataset, and  62.6 mAP (2D map) / 61.4 mAP (3D map) on Argoverse2 dataset.
Extending  MapTRv2 to centerline paves the way for end-to-end planning.

\subsection{Qualitative Results}

\begin{figure*}[htbp!]
    \centering
    \includegraphics[width=0.9\linewidth]{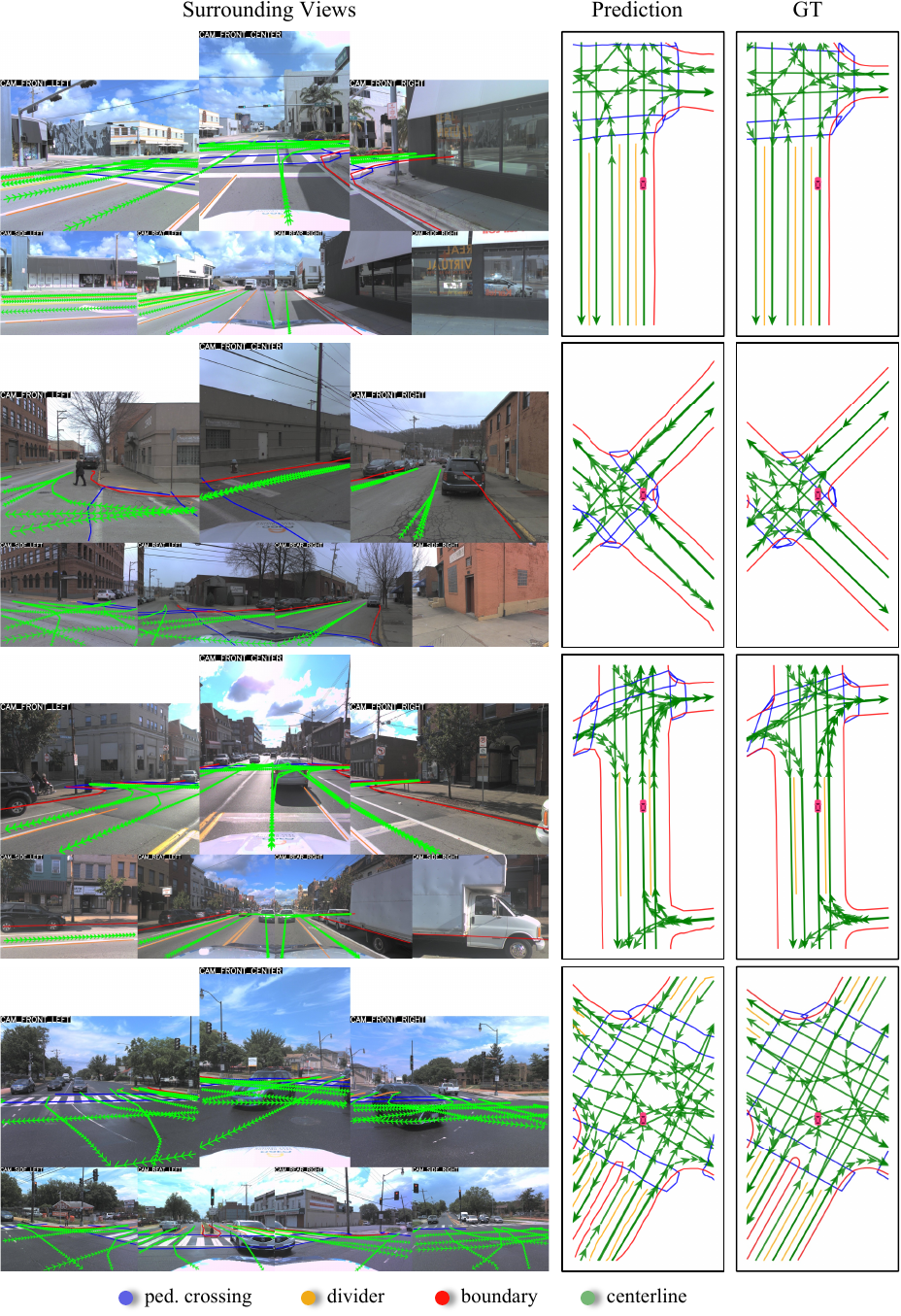}
    \caption{Qualitative results on Argoverse2 \texttt{val} set. Since we predict 3D vectorized HD map, the predictions can be accurately rendered on the surrounding view images.}
    \label{fig:av2}
\end{figure*}

\begin{figure*}[htbp!]
    \centering
    \includegraphics[width=0.9\linewidth]{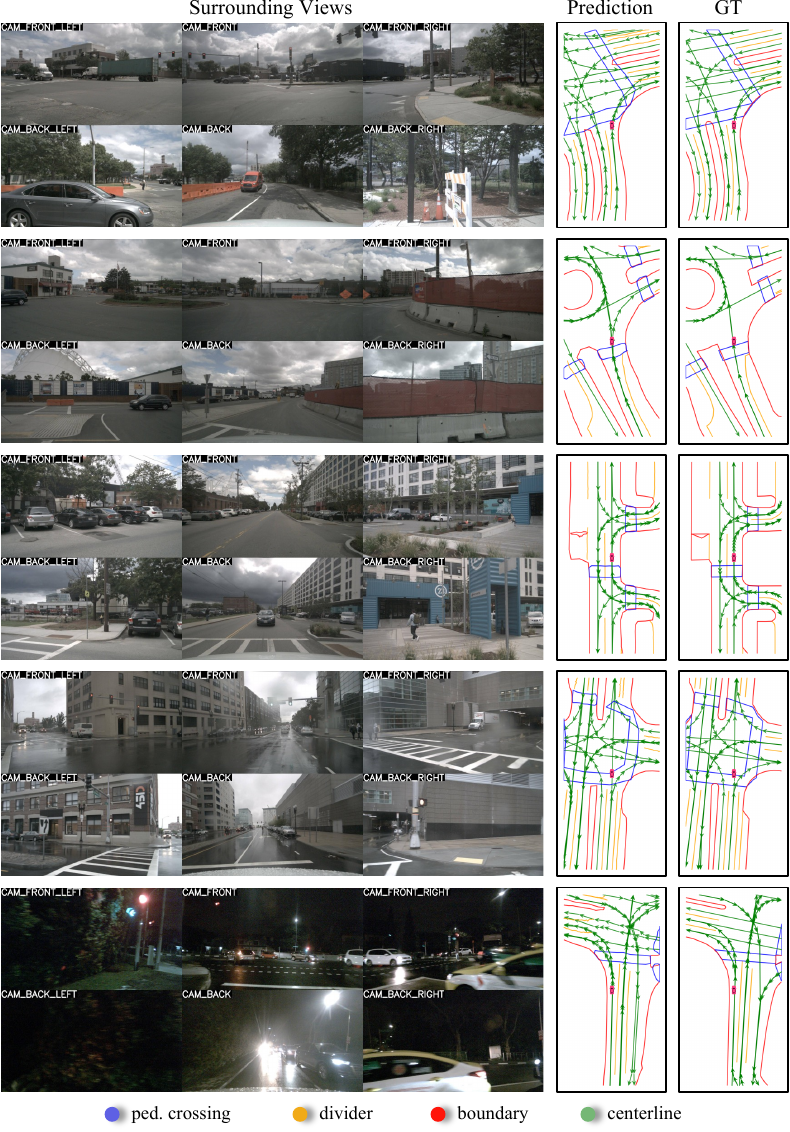}
    \caption{Qualitative results on nuScenes \texttt{val} set.}
    \label{fig:nusc}
\end{figure*}

We show the predicted vectorized HD map results of Argoverse2 and nuScenes datasets respectively in Fig.~\ref{fig:av2} and Fig.~\ref{fig:nusc}. MapTRv2 maintains stable and impressive results in complex and various driving scenes.  More qualitative results are available at the project website \url{https://github.com/hustvl/MapTR}.

\section{Conclusion}
MapTRv2 is a structured end-to-end framework for efficient online vectorized HD map construction, 
which  adopts a simple encoder-decoder Transformer architecture and hierarchical bipartite matching to perform map element learning based on the proposed permutation-equivalent modeling.
Extensive experiments show that the proposed method can precisely perceive  map elements of arbitrary shape in both nuScenes and Argoverse2 datasets. We hope MapTRv2 can serve as a basic module of self-driving system and boost the development of downstream tasks (\eg, motion prediction and planning).

\backmatter




\section*{Declarations}


\begin{itemize}
\item \textbf{Conflict of Interest.} All authors declare no conflicts of interest.
\item \textbf{Data availability.} The data that supports the findings of this study are publicly available at nuScenes~\cite{nuscenes} and Argoverse2~\cite{av2}.
\item \textbf{Code availability.} The toolkit and experimental results are publicly available at \url{https://github.com/hustvl/MapTR}. 
\end{itemize}

\bibliography{sn-bibliography}

\end{document}